%% file: root.tex
\documentclass[journal,twoside]{IEEEtran}

\usepackage{cite}

\ifCLASSINFOpdf
  \usepackage[pdftex]{graphicx}
\else
  \usepackage[dvips]{graphicx}
\fi
\usepackage{placeins}

\usepackage{amsmath}
\usepackage{amssymb}  % assumes amsmath package
\usepackage{mathtools}

\usepackage{algorithm} 
\usepackage{algpseudocode} 
\usepackage{array}
\usepackage[caption=false,font=footnotesize]{subfig}
\usepackage{url}
\usepackage{siunitx}
\usepackage{hyperref}
\usepackage{multirow}
\usepackage[multiple]{footmisc}

\usepackage{color}
\newcommand{\new}[1]{{#1}}

\input{chapters/mathdef}

\begin{document}
\title{Perceptive Locomotion through Nonlinear Model Predictive Control}

\author{Ruben~Grandia$^1$,
        Fabian~Jenelten$^1$,
        Shaohui~Yang$^2$,
        Farbod~Farshidian$^1$,
        and~Marco~Hutter$^1$% <-this % stops a space 
\thanks{This research was supported by the Swiss National Science Foundation (SNSF) as part of project No.188596 and by the Swiss National Science Foundation through the National Centre of Competence in Research Robotics (NCCR
Robotics). This project has received funding from the European Union’s Horizon 2020 research and innovation programme under grant agreement No 780883.}% <-this % stops a space
\thanks{$^1$R. Grandia, F. Jenelten, F. Farshidian and M. Hutter are with the Department of Mechanical and Process Engineering, ETH Zurich, Switzerland. {\tt\small \{rgrandia,fabianje,farbodf,mahutter\}@ethz.ch}.}% <-this % stops a space
\thanks{$^2$S. Yang is with the Automatic Control Laboratory, École polytechnique fédérale de Lausanne (EPFL), Switzerland. {\tt\small shaohui.yang@epfl.ch}.}% <-this % stops a space
\thanks{Manuscript received March 13, 2022; revised August 17, 2022.}}

% The paper headers, name and title will appear is for the odd numbered pages
%\markboth{IEEE TRANSACTIONS ON ROBOTICS,~Vol.~XX, No.~X, AUGUST~2022}%
%{Grandia \MakeLowercase{\textit{et al.}}: Perceptive Locomotion through Nonlinear Model Predictive Control}

\maketitle

\input{chapters/abstract}

% %%%%%%%%%% Main text %%%%%%%%%%%%%%%%%%%%%%%

\input{chapters/main}

\appendices
\input{chapters/appendix}

\input{chapters/acknowledgement}

% %%%%%%%%%% Bibliography %%%%%%%%%%%%%%%%%%%%%%%

\ifCLASSOPTIONcaptionsoff
  \newpage
\fi


\typeout{}
\bibliographystyle{bibtex/IEEEtran}
\bibliography{bibtex/IEEEabrv,bibtex/library.bib}

% %%%%%%%%%% Biography %%%%%%%%%%%%%%%%%%%%%%%
\input{chapters/biography}

\end{document}

%% file: chapters/mathdef.tex
\newcommand\shortdots{\makebox[1em][c]{.\hss.\hss.}\thinspace}

\newcommand{\dt}{\textnormal{d}t}

\newcommand{\argmin}{\mathop{\mathrm{argmin}}}

\newcommand{\R}{\mathbb{R}}

\newcommand{\vep}{\mbox{\boldmath $\epsilon$}}

\newcommand{\vth}{\mbox{\boldmath $\theta$}}

\newcommand{\vlambda}{\mbox{\boldmath $\lambda$}}
\newcommand{\vmu}{\mbox{\boldmath $\mu$}}
\newcommand{\vnu}{\boldsymbol{ \nu}}

\newcommand{\vtau}{\mbox{\boldmath $\tau$}}

\newcommand{\vom}{\mbox{\boldmath $\omega$}}

\newcommand{\vSi}{\mathbf \Sigma}

\newcommand{\vb}{\mathbf b}
\newcommand{\vc}{\mathbf c}

\newcommand{\vf}{\mathbf f}
\newcommand{\vg}{\mathbf g}
\newcommand{\vh}{\mathbf h}

\newcommand{\vk}{\mathbf k}

\newcommand{\vn}{\mathbf n}

\newcommand{\vp}{\mathbf p}
\newcommand{\vq}{\mathbf q}

\newcommand{\vs}{\mathbf s}

\newcommand{\vu}{\mathbf u}
\newcommand{\vv}{\mathbf v}
\newcommand{\vw}{\mathbf w}
\newcommand{\vx}{\mathbf x}

\newcommand{\zero}{\mathbf 0}

\newcommand{\vA}{\mathbf A}
\newcommand{\vB}{\mathbf B}
\newcommand{\vC}{\mathbf C}
\newcommand{\vD}{\mathbf D}

\newcommand{\vF}{\mathbf F}
\newcommand{\vG}{\mathbf G}
\newcommand{\vH}{\mathbf H}
\newcommand{\vI}{\mathbf I}
\newcommand{\vJ}{\mathbf J}
\newcommand{\vK}{\mathbf K}

\newcommand{\vM}{\mathbf M}

\newcommand{\vP}{\mathbf P}

\newcommand{\vR}{\mathbf R}
\newcommand{\vS}{\mathbf S}
\newcommand{\vT}{\mathbf T}
\newcommand{\vU}{\mathbf U}

\newcommand{\vW}{\mathbf W}
\newcommand{\vX}{\mathbf X}

%% file: chapters/abstract.tex
\begin{abstract}
Dynamic locomotion in rough terrain requires accurate foot placement, collision avoidance, and planning of the underactuated dynamics of the system. 
Reliably optimizing for such motions and interactions in the presence of imperfect and often incomplete perceptive information is challenging.
\new{We present a complete perception, planning, and control pipeline, that can optimize motions for all degrees of freedom of the robot in real-time.
To mitigate the numerical challenges posed by the terrain a sequence of convex inequality constraints is extracted as local approximations of foothold feasibility and embedded into an online model predictive controller.
Steppability classification, plane segmentation, and a signed distance field are precomputed per elevation map to minimize the computational effort during the optimization.
A combination of multiple-shooting, real-time iteration, and a filter-based line-search are used to solve the formulated problem reliably and at high rate.}
We validate the proposed method in scenarios with gaps, slopes, and stepping stones in simulation and experimentally on the ANYmal quadruped platform, resulting in state-of-the-art dynamic climbing.
\end{abstract}


%% file: chapters/main.tex
\nocite{video}
\section{Introduction}
\FloatBarrier

\begin{figure}[!t]
\centering
\includegraphics[width=\columnwidth,
trim={75 205 0 0},clip]{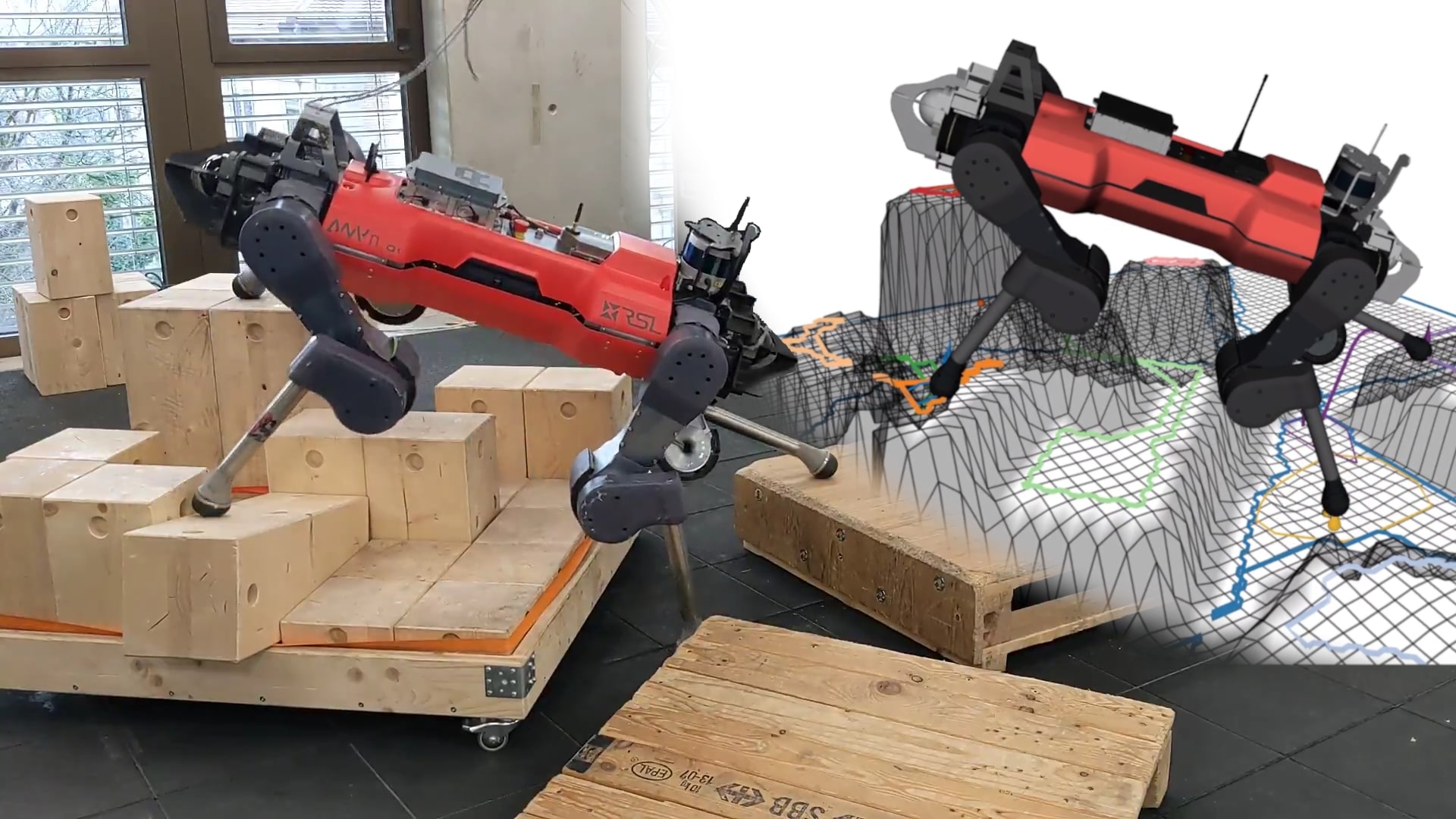}
\caption{ANYmal walking on uneven stepping stones. In the shown configuration, the top foothold is \SI{60}{\centi\meter} above the lowest foothold. The top right visualizes the internal terrain representation used by the controller.}
\label{fig:perc:ANYmal}
\end{figure}

\IEEEPARstart{I}{nspired} by nature, the field of legged robotics aims to enable the deployment of autonomous systems in rough and complex environments. Indeed, during the recent DARPA subterranean challenge, legged robots were widely adopted, and highly successful \cite{tranzatto2021cerberus,bouman2020autonomous}. Still, complex terrains that require precise foot placements, e.g., negative obstacles and stepping stones as shown in Fig.~\ref{fig:perc:ANYmal}, remain difficult.

A key challenge lies in the fact that both the terrain and the system dynamics impose constraints on contact location, force, and timing. When taking a model-based approach, mature methods exist for perceptive locomotion with a slow, static gait~\cite{kalakrishnan2010fast,belter2016adaptive,mastalli2020motion,fankhauser2018robust,griffin2019footstep} and for blind, dynamic locomotion that assumes flat terrain~\cite{bellicoso2018dynamic,bledt2017policy,di2018dynamic}. Learning-based controllers have recently shown the ability to generalize blind locomotion to challenging terrain with incredible robustness~\cite{lee2020learning,siekmann2021blind,miki2022learning}. Still, tightly integrating perception to achieve coordinated and precise foot placement remains an active research problem. 

%This challenge has led to methods that combine model-based planning and learning-based control~\cite{lowrey2018,tsounis2020deepgait,gangapurwala2020rloc,gangapurwala2021real}. In such a setup, model-based planning plays a crucial role in discovering complex movements that require precise control over long time horizons.

In an effort to extend dynamic locomotion to uneven terrain, several methods have been proposed to augment foothold selection algorithms with perceptive information~\cite{jenelten2020perceptive,kim2020vision,villarreal2020mpc}. These approaches build on a strict hierarchy of first selecting footholds and optimizing torso motion afterward. This decomposition reduces the computational complexity but relies on hand-crafted coordination between the two modules. Additionally, separating the legs from the torso optimization makes it difficult to consider kinematic limits and collision avoidance between limbs and terrain.

Trajectory optimization \new{where torso and leg motions are jointly optimized} has shown impressive results in simulation~\cite{mordatch2012discovery,winkler2018gait,dai2014wholebody} and removes the need for engineered torso-foot coordination. Complex motions can be automatically discovered by including the entire terrain in the optimization. However, computation times are often too long for online deployment. Additionally, due to the non-convexity, non-linearity, and discontinuity introduced by optimizing over arbitrary terrain, these methods can get stuck in poor local minima. Dedicated work on providing an initial guess is needed to find feasible motions reliably~\cite{melon2020reliable}. 

This work presents a planning and control framework that optimizes over all degrees of freedom of the robot, considers collision avoidance with the terrain, and enables complex dynamic maneuvers in rough terrain. The method is centered around nonlinear Model Predictive Control (MPC) with a multiple-shooting discretization~\cite{bock1984multiple,rawlings2017model}. However, in contrast to the aforementioned work, where the full terrain is integrated into the optimization, we get a handle on the numerical difficulty introduced by the terrain by exposing the terrain as a series of geometric primitives that approximate the local terrain. In this case, we use convex polygons as foot placement constraints, but different shapes can be used as long as they lead to well-posed constraints in the optimization. Additionally, a signed distance field (SDF) is used for collision avoidance. We empirically demonstrate that such a strategy is an excellent trade-off between giving freedom to the optimization to discover complex motions and the reliability with which we can solve the formulated problem.

\subsection{Contributions}
We present a novel approach to locomotion in challenging terrain where perceptive information needs to be considered and nontrivial motions are required. The complete perception, planning, and control pipeline contains the following contributions:

\begin{itemize}
	\item \new{We perform simultaneous and real-time optimization of all degrees of freedom of the robot for dynamic motions across rough terrain. Perceptive information is encoded through a sequence of geometric primitives that capture local foothold constraints and a signed distance field used for collision avoidance. 
	}
    \item \new{The proposed combination of a multiple-shooting transcription, sequential quadratic programming, and a filter-based line-search enables fast and reliable online solutions to the nonlinear optimal control problem.}
    \item \new{We provide a detailed description of the implemented MPC problem, its integration with whole-body and reactive control modules, and extensive experimental validation of the resulting locomotion controller.}
\end{itemize}

\new{The MPC implementation is publicly available as part of the OCS2 toolbox\footnote{\href{https://github.com/leggedrobotics/ocs2/tree/main/ocs2\_sqp}{https://github.com/leggedrobotics/ocs2}} \cite{OCS2}. The implemented online segmentation of the elevation map, and the efficient precomputation of a signed distance field are contributed to existing open-source repositories%
\footnote{\new{\href{https://github.com/leggedrobotics/elevation\_mapping\_cupy/tree/main/plane\_segmentation}{https://github.com/leggedrobotics/elevation\_mapping\_cupy}}}%
\footnote{\new{\href{https://github.com/ANYbotics/grid\_map/tree/master/grid\_map\_sdf}{https://github.com/ANYbotics/grid\_map}}}.}

\subsection{Outline}
\begin{figure}[!t]
\centering
\includegraphics[width=0.95\columnwidth]{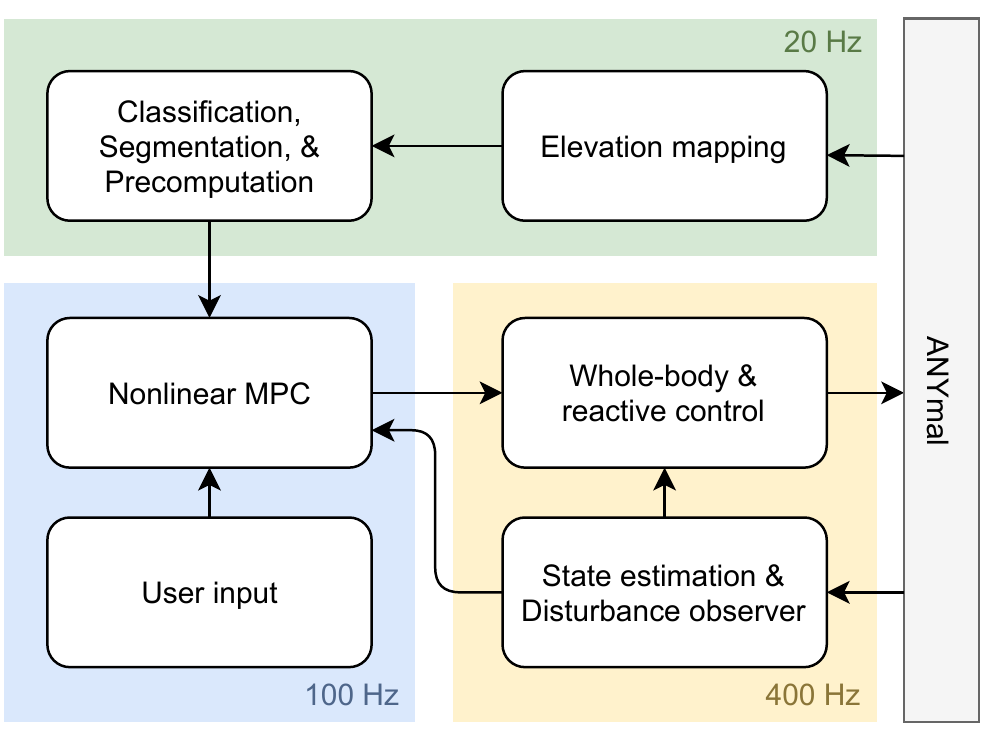}
\caption{Schematic overview of the proposed method together with the update rate of each component.}
\label{fig:perc:method_overview}
\end{figure}

An overview of the proposed method is given in Fig.~\ref{fig:perc:method_overview}. The perception pipeline at the top of the diagram runs at \SI{20}{\hertz} and is based on an elevation map constructed from pointcloud information. For each map update, classification, segmentation, and other precomputation are performed to prepare for the high number of perceptive queries during motion optimization. At the core of the framework, we use nonlinear MPC at \SI{100}{\hertz} to plan a motion for all degrees of freedom and bring together user input, perceptive information, and the measured state of the robot. Finally, state estimation, whole-body torque control, and reactive behaviors are executed at a rate of \SI{400}{\hertz}.

After a review of related work in section~\ref{sect:perc:related_work}, this paper is structured similarly to Fig.~\ref{fig:perc:method_overview}. First, we present the perception pipeline in section \ref{sect:perc:perception}. Afterward, the formulated optimal control problem and corresponding numerical optimization strategy are discussed in sections~\ref{sect:perc:motion_optimization}~{\&}~\ref{sect:perc:numerical_optimization}. We introduce the motion execution layer in section \ref{sect:perc:motion_execution}. The resulting method is evaluated on the quadrupedal robot ANYmal~\cite{hutter2016anymal} (see Fig.~\ref{fig:perc:ANYmal}) in section~\ref{sect:perc:results}, and concluded with
section~\ref{sect:perc:conclusion}.

\section{Related Work}
\label{sect:perc:related_work}

\subsection{Decomposing locomotion}
When assuming a quasi-static gait with a predetermined stepping sequence, the planning problem on rough terrain can be simplified and decomposed into individual contact transitions, as demonstrated in the early work on \textit{LittleDog} \cite{kolter2008control,kalakrishnan2010fast}. In a one-step-ahead fashion, one can check the next foothold for kinematic feasibility, feasibility w.r.t. the terrain, and the existence of a statically stable transition. This problem can be efficiently solved by sampling and checking candidate footholds \cite{tonneau2018efficient}. Afterward, a collision-free swing leg trajectory to the desired foothold can be generated with CHOMP \cite{zucker2013chomp} based on an SDF. Fully onboard perception and control with such an approach were achieved by Fankhauser et al.~\cite{fankhauser2018robust}. Instead of one-step-ahead planning, an RRT graph can be built to plan further ahead \cite{belter2016adaptive}. Sampling over templated foothold transitions achieves similar results~\cite{mastalli2015online,mastalli2020motion}.

In this work, we turn our attention to dynamic gaits, where statically stable transitions between contact configurations are not available. In model-based approaches to dynamic, perceptive locomotion, a distinction can be made between methods where the footholds locations are determined separately from the torso and those where the foothold locations and torso motions are jointly optimized. 

Several methods in which footholds are selected before optimizing the torso motions, initially designed for flat terrain, have been adapted to traverse rough terrain~\cite{bajracharya2013high,bazeille2014quadruped}. These methods typically employ some form of Raibert heuristic \cite{raibert1986legged} to select the next foothold and adapt it based on perceptive information such as a traversability estimate~\cite{wermelinger2016navigation}. The work of Bellicoso et al.~\cite{bellicoso2018dynamic} was extended by including a batch search for feasible footholds based on a given terrain map and foothold scoring \cite{jenelten2020perceptive}. Similarly, in~\cite{kim2020vision}, the foot placement is adapted based on visual information resulting in dynamic trotting and jumping motions. In~\cite{magana2019fast}, the authors proposed to train a convolutional neural network (CNN) to speed up the online evaluation of such a foothold adaptation pipeline. This CNN was combined with the MPC strategy in~\cite{di2018dynamic} to achieve perceptive locomotion in simulation~\cite{villarreal2020mpc}. In~\cite{gangapurwala2022rloc}~and~\cite{yu2021visuallocomotion}, a Reinforcement Learning (RL) policy has replaced the heuristic foothold selection.

However, since foothold locations are chosen before optimizing the torso motion, their effect on dynamic stability and kinematic feasibility is not directly considered, requiring additional heuristics to coordinate feet and torso motions to satisfy whole-body kinematics and dynamics. Moreover, it becomes hard to consider collisions of the leg with the terrain because the foothold is already fixed. In our approach, we use the same heuristics to find a suitable nominal foothold in the terrain. However, instead of fixing the foothold to that particular location, a region is extracted around the heuristic in which the foothold is allowed to be optimized.

The benefit of jointly optimizing torso and leg motions has been demonstrated in the field of trajectory optimization. One of the first demonstrations of simultaneous optimization of foot placement and a zero-moment point (ZMP) \cite{vukobratovic2004zeromoment} trajectory was achieved by adding 2D foot locations as decision variables to an MPC algorithm~\cite{herdt2010online}. More recently, Kinodynamic \cite{farshidian2017efficient}, Centroidal~\cite{orin2013centroidal,sleiman2021unified}, and full dynamics models \cite{pardo2017hybrid,herzog2016structured} have been used for simultaneous optimization of 3D foot locations and body motion. Alternatively, a single rigid body dynamics (SRBD) model \new{or other simplified torso models} can be extended with decision variables for Cartesian foothold locations~\cite{winkler2018gait}\new{,\cite{jenelten2021TAMOLS}}. Real-time capable methods have been proposed with the specification of leg motions on position~\cite{bledt2017policy}, velocity~\cite{farshidian2017realtime}, or acceleration level~\cite{neunert2018wholebody}. One challenge of this line of work is the computational complexity arising from the high dimensional models, already in the case of locomotion on flat terrain. Our method also uses a high-dimensional model and falls in this category. A key consideration when extending the formulations with perceptive information has thus been to keep computation within real-time constraints.

Finally, several methods exist that additionally optimize gait timings or even the contact sequence together with the whole-body motion. This can be achieved through complementarity constraints~\cite{mordatch2012discovery,posa2014direct,dai2014wholebody}, mixed-integer programming\cite{aceituno2018simultaneous,marcucci2017approximate}, or by explicitly integrating contact models into the optimization \cite{neunert2018wholebody,carius2018trajectory}. Alternatively, the duration of each contact phase can be included as a decision variable~\cite{ponton2018ontime,winkler2018gait} or found through bilevel optimization~\cite{farshidian2017sequential,seyde2019locomotion}. \new{ 
However, such methods are prone to poor local optima and reliably solving the optimization problems in real-time remains challenging.}

\subsection{Terrain representation}
The use of an elevation map has a long-standing history in the field of legged robotics~\cite{herbert1989terrain}, and it is still an integral part of many perceptive locomotion controllers today. Approaches where footholds are selected based on a local search or sampling-based algorithm can directly operate on such a structure. However, more work is needed when integrating the terrain into a gradient-based optimization. 

Winkler et al.~\cite{winkler2018gait} uses an elevation map for both foot placement and collision avoidance. The splines representing the foot motion are constrained to start and end on the terrain with equality constraints. An inequality constraint is used to avoid the terrain in the middle of the swing phase. Ignoring the discontinuity and non-convexity from the terrain makes this approach prone to poor local minima, motivating specialized initialization schemes~\cite{melon2020reliable} for this framework.

In~\cite{jenelten2021TAMOLS}, a graduated optimization scheme is used, where a first optimization is carried out over a smoothened version of the terrain. The solution of this first optimization is then used to initialize an optimization over the actual elevation map. In a similar spirit, Mordatch \cite{mordatch2012discovery} considers a general 3D environment and uses a soft-min operator to smoothen the closest point computation. A continuation scheme is used to gradually increase the difficulty of the problem over consecutive optimizations.

Deits et al.~\cite{deits2014footstep} describe a planning approach over rough terrain based on mixed-integer quadratic programming (MIQP). Similar to \cite{griffin2019footstep}, convex safe regions are extracted from the terrain, and footstep assignment to a region is formulated as a discrete decision. The foothold optimization is simplified because only convex, safe regions are considered during planning. \new{Furthermore, the implementation relied on manual seeding of convex regions by a human operator}. We follow the same philosophy of presenting the terrain as a convex region to the optimization. However, we remove the mixed-integer aspect by pre-selecting the convex region. The benefits are two-fold: First, we do not require a global convex decomposition of the terrain, which is a hard problem in general \cite{bertrand2020detecting}, and instead, only produce a local convex region centered around a nominal foothold. Second, the MIQP approach does not allow for nonlinear costs and dynamics, which limits the range of motions that can be expressed. We first explored the proposed terrain representation as part of our previous work~\cite{grandia2021multi}, but relied on offline mapping, manual terrain segmentation, and did not yet consider terrain collisions. \new{In~\cite{bjelonic2022offline}, we applied this idea to wheeled-legged robots, but again relied on offline mapping and segmentation. Moreover, as discussed in the next section, in both \cite{grandia2021multi} and \cite{bjelonic2022offline}, we used a different solver, which was found to be insufficient for the scenarios in this work.} 

\begin{figure*}[!t]
\centering
\includegraphics[width=\linewidth]{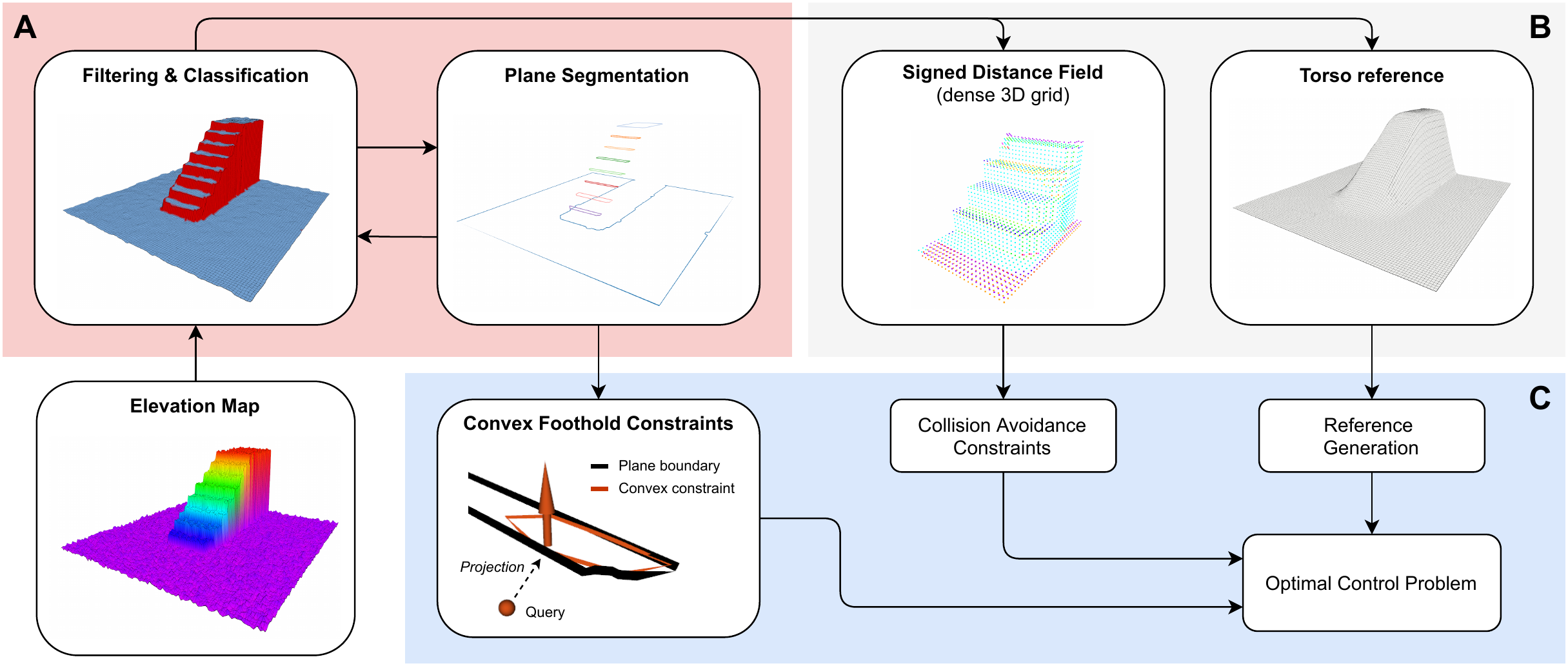}
\caption{Perception pipeline overview. (\textbf{A}) The elevation map is filtered and classified into steppable and non-steppable cells \new{[Section~\ref{sect:perc:filtering_and_classification}]}. All steppable areas are segmented into planes \new{[Section~\ref{sect:perc:plane_segmentation}]}. After segmentation, the steppablity classification is refined. (\textbf{B}) A signed distance field \new{[Section~\ref{sect:perc:signed_distance_field}]} and torso reference layer \new{[Section~\ref{sect:perc:torso_reference_layer}]} are precomputed to reduce the required computation time during optimization. \textbf{(C)} Convex foothold constraints in \new{\eqref{eq:perc:foothold_position_constraint}} are obtained from the plane segmentation. The signed distance field enables collision avoidance \new{in \eqref{eq:perc:sdf_inequality}}, and the torso reference is used to generate height and orientation references \new{[Section~\ref{sect:perc:reference_generation}]}.}
\label{fig:perc:perception_overview}
\end{figure*}

\subsection{Motion Optimization}
For trajectory optimization, large-scale optimization software like SNOPT~\cite{gill2005snopt} and IPOPT~\cite{wachter2006implementation} are popular. They are the workhorse for offline trajectory optimization in the work of Winkler \cite{winkler2018gait}, Dai \cite{dai2014wholebody}, Mordatch \cite{mordatch2012discovery}, Posa \cite{posa2014direct}, and Pardo \cite{pardo2017hybrid}. These works show a great range of motions in simulation, but it typically takes minutes to hours to find a solution.

A different line of work uses specialized solvers that exploit the sparsity that arises from a sequential decision making process. Several variants of Differential Dynamic Programming (DDP)~\cite{jacobson1970differential} have been proposed in the context of robotic motion optimization, e.g., iLQR~\cite{tassa2012synthesis,howell2019ALTRO}, SLQ~\cite{farshidian2017efficient}, and FDDP~\cite{mastalli2020crocoddyl}. 

With a slightly different view on the problem, the field of (nonlinear) model predictive control~\cite{mayne2014model,rawlings2017model} has specialized in solving successive optimal control problems under real-time constraints. See \cite{kouzoupis2018recent} for a comparison of state-of-the-art quadratic programming (QP) solvers that form the core of second-order optimization approaches to the nonlinear problem. For time-critical applications, the real-time iteration scheme can be used to trade optimality for lower computational demands~\cite{diehl2005realtime}: In a Sequential Quadratic Programming (SQP) approach to the nonlinear problem, at each control instance, only a single QP optimization step is performed.

The current work was initially built on top of a solver in the first category~\cite{farshidian2017efficient}. However, a significant risk in classical DDP-based approaches is the need to perform a nonlinear system rollout along the entire horizon. Despite the use of a feedback policy, these forward rollouts can diverge, especially in the presence of underactuated dynamics. This same observation motivated Mastalli et al.\ to design FDDP to maintain \textit{gaps} between shorter rollouts, resulting in a formulation that is equivalent to direct multiple-shooting formulations with only equality constraints~\cite{mastalli2020crocoddyl,bock1984multiple}. \new{Giftthaler et al.\ \cite{giftthaler2018family} studied several combinations of iLQR and multiple-shooting but did not yet consider constraints beyond system dynamics nor a line-search procedure to determine the stepsize. Furthermore, experiments were limited to simple, flat terrain walking.}

We directly follow the multiple-shooting approach with a real-time iteration scheme and leverage the efficient structure exploiting QP solver HPIPM~\cite{frison2020hpipm}. However, as also mentioned in \new{both}~\cite{mastalli2020crocoddyl} \new{and \cite{giftthaler2018family}}, one difficulty is posed in deciding a stepsize for nonlinear problems, where one now has to monitor both the violation of the system dynamics and minimization of the cost function. To prevent an arbitrary trade-off through a merit function, we suggest using a filter-based line-search instead~\cite{fletcher2002nonlinear}, which allows a step to be accepted if it reduces either the objective function or the constraint violation. \new{As we will demonstrate in the result section, these choices contribute to the robustness of the solver in challenging scenarios.}

\section{Terrain perception and segmentation}
\label{sect:perc:perception}

An overview of the perception pipeline and its relation to the MPC controller is provided in Fig.~\ref{fig:perc:perception_overview}. The pipeline can be divided into three parts: (\textbf{A}) steppability classification and segmentation, (\textbf{B}) precomputation of the SDF and torso reference, and (\textbf{C}) integration into the optimal control problem.

The elevation map, represented as a 2.5D grid \cite{fankhauser2016universal} with a \SI{4}{\centi\meter} resolution is provided by the GPU based implementation introduced in \cite{miki2022elevation}. The subsequent map processing presented in this work runs on the CPU and is made available as part of that same open-source library. Both (\textbf{A}) and (\textbf{B}) are computed once per map and run \new{at \SI{20}{\hertz},} asynchronously to the motion optimization in (\textbf{C}).

\subsection{Filtering \& Classification}\label{sect:perc:filtering_and_classification}
The provided elevation map contains empty cells in occluded areas. As a first step, we perform \textit{inpainting} by filling each cell with the minimum value found along the occlusion border. Afterwards, a median filter is used to reduce noise and outliers in the map. 

Steppablity classification is performed by thresholding the local surface inclination and the local roughness estimated through the standard deviation~\cite{chilian2009stereo}. Both quantities can be computed with a single pass through the considered neighbourhood of size $N$:
\begin{equation}
    \vmu = \frac{1}{N} \sum_i \vc_i, \quad  \vS = \frac{1}{N} \sum_i \vc_i \vc^\top_i, \quad  \vSi =  \vS - \vmu \vmu^\top,
    \label{eq:perc:covariance}
\end{equation}
where $\vmu$ and $\vS$ are the first and second moment, and $\vSi \in \R^{3\times3}$ is the positive semi-definite covariance matrix of the cell positions $\vc_i$. The variance in normal direction, $\sigma_n^2$, is then the smallest eigenvalue of $\vSi$, and the surface normal, $\vn$, is the corresponding eigenvector. For steppability classification we use a neighbourhood of $N=9$, and set a threshold of \SI{2}{\centi\meter} on the standard deviation in normal direction and a maximum inclination of \SI{35}{\degree}, resulting in the following classification:
\begin{equation}
    \text{steppability} = \begin{cases}
1  & \text{if } \sigma_n \leq 0.02, \text{ and } n_z \geq 0.82, \\
0            & \text{otherwise},
\end{cases}
\label{eq:perc:steppability}
\end{equation}
where $n_z$ denotes the z-coordinate of the surface normal. 

\subsection{Plane Segmentation}
\label{sect:perc:plane_segmentation}
After the initial classification, the plane segmentation starts by identifying continuous regions with the help of a connected component labelling \cite{wu2009optimizing}. For each connected region of cells, we compute again the covariance as in \eqref{eq:perc:covariance}, where $N$ is now the number of cells in the connected region, and accept the region as a plane based on the following criteria:
\begin{equation}
    \text{planarity} = \begin{cases}
1  & \text{if } \sigma_n \leq 0.025, n_z \geq 0.87, \text{ and } N \geq 4 \\
0            & \text{otherwise}.
\end{cases}
\label{eq:perc:planarity}
\end{equation}
Notice that here we loosen the bound on the standard deviation to \SI{2.5}{cm}, tighten the bound on the inclination to \SI{30}{\degree}, and add the constraint that at least 4 cells form a region. 
% Offsetting the thresholds between \eqref{eq:perc:planarity} and \eqref{eq:perc:steppability} acts as a deadzone and prevents over-segmentation in case several cells are right around the threshold in \eqref{eq:perc:steppability}. 
If the planarity condition is met, the surface normal and mean of the points define the plane.

If a region fails the planarity condition, we trigger RANSAC \cite{schnabel2007efficient} on that subset of the data. The same criteria in \eqref{eq:perc:planarity} are used to find smaller planes within the connected region. After the algorithm terminates, all cells that have not been included in any plane have their steppability updated and set to 0.  

\begin{figure}[!tb]
    \centering
    \begin{minipage}{.32\columnwidth}
        \centering
        \includegraphics[width=0.95\linewidth]{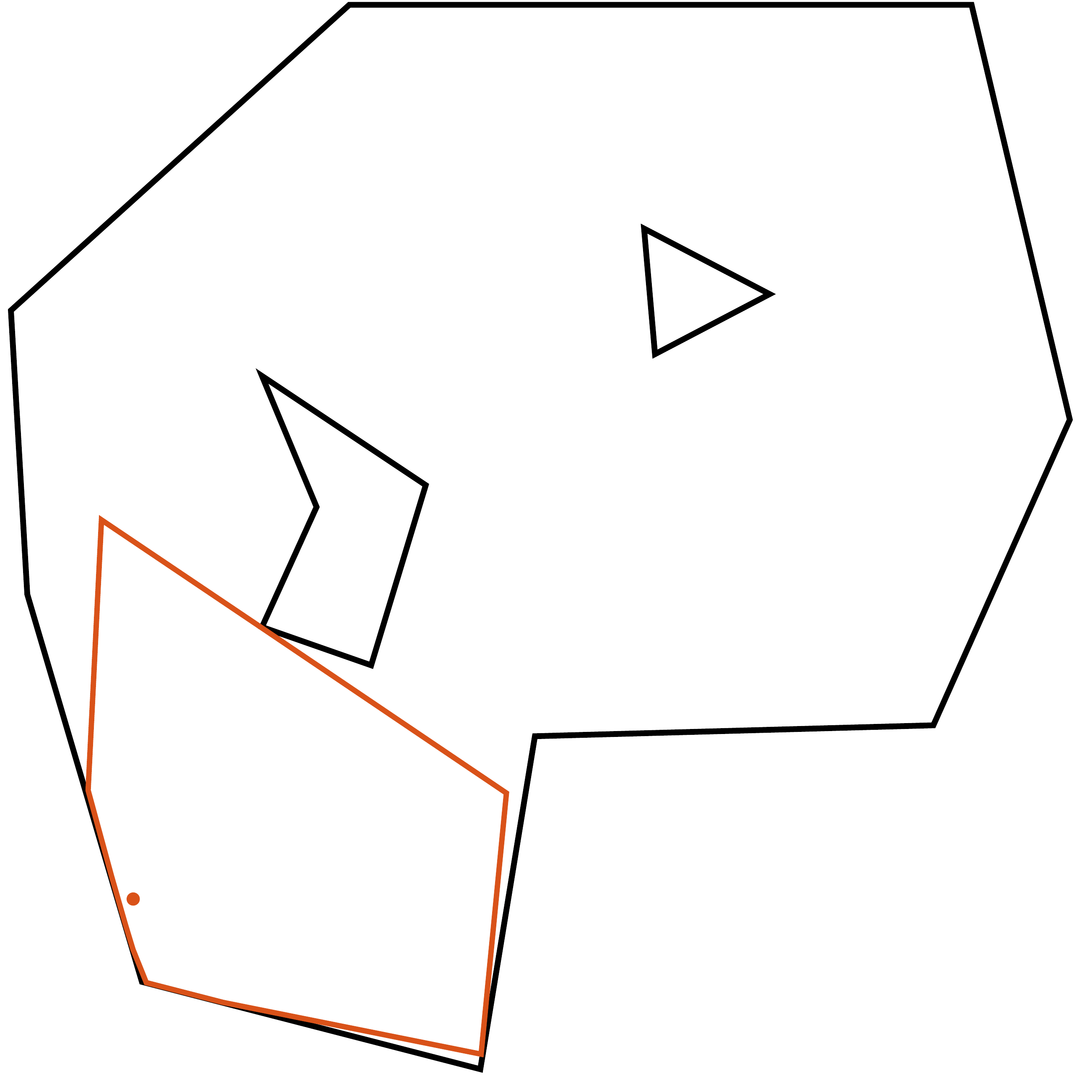}
    \end{minipage}
    \begin{minipage}{.32\columnwidth}
        \centering
        \includegraphics[width=0.95\linewidth]{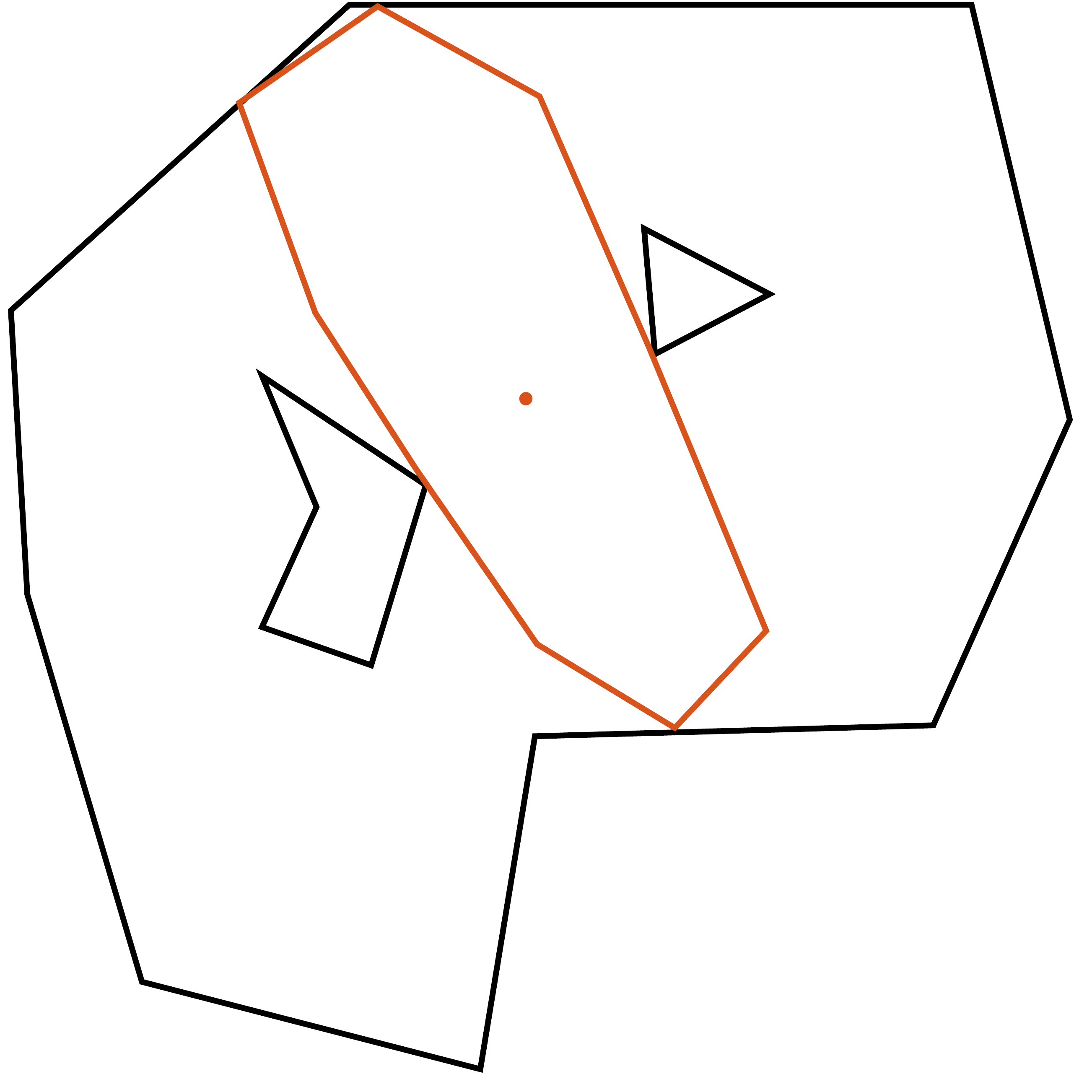}
    \end{minipage}
        \begin{minipage}{.32\columnwidth}
        \centering
        \includegraphics[width=0.95\linewidth]{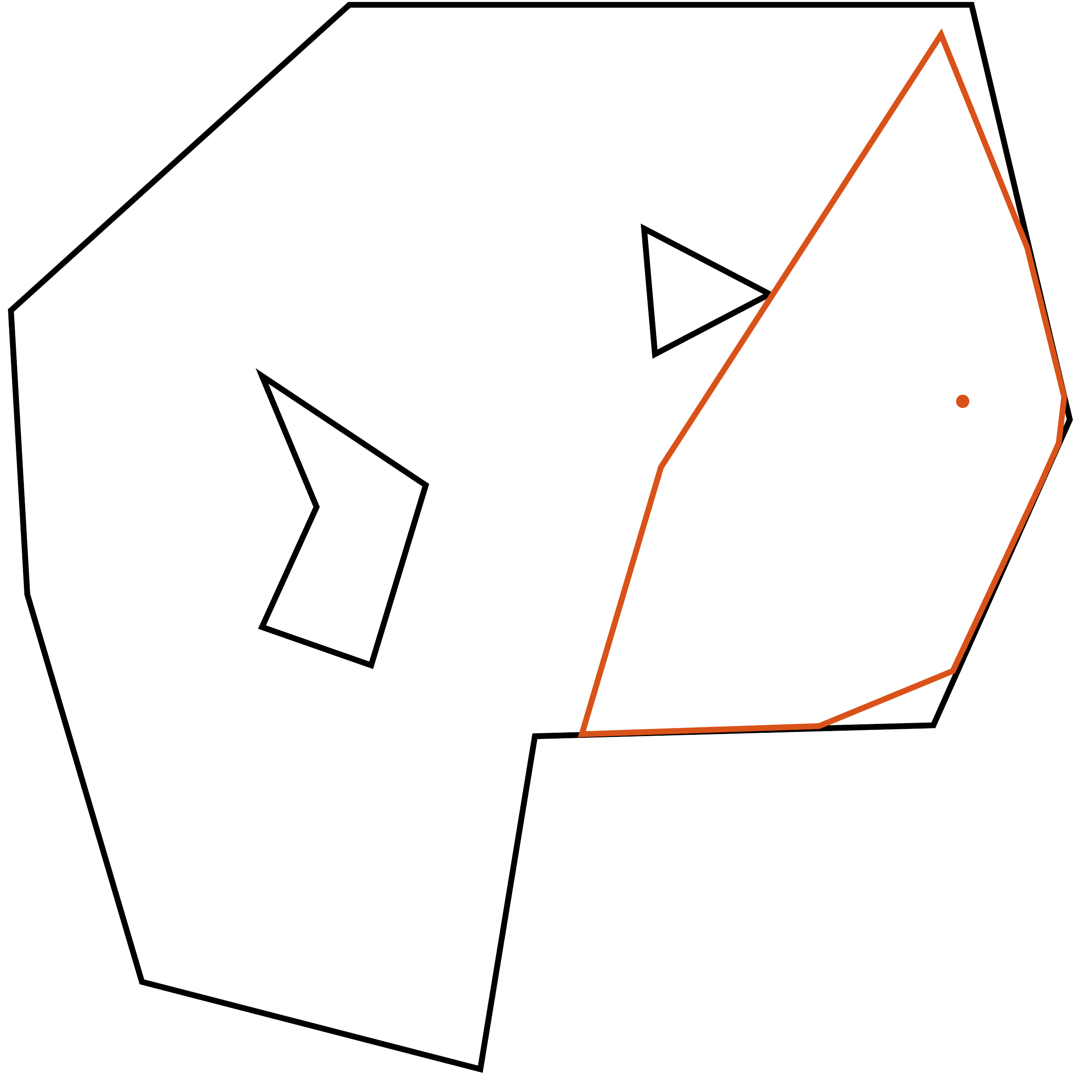}
    \end{minipage}
    \caption{\new{An example of a segmented region represented by a non-convex outer polygon and two non-overlapping holes (drawn in black). Three different local convex approximations (drawn in orange) are shown that are found around query points with the iterative algorithm described in section~\ref{sect:perc:cost_definition}.}}
    \label{fig:perc:polygon_examples}
\end{figure}

At this point, we have a set of plane parameters with connected regions of the map assigned to them. For each of these regions, we now extract a 2D contour from the elevation map \cite{suzuki1985topological}, and project it along the z-axis to the plane to define the boundary in the frame of the plane. It is important to consider that regions can have holes, \new{for example, when a free-standing obstacle is located in the middle of an open floor. The boundary of each segmented region is therefore represented by} an outer polygon together with a set of polygons that trace enclosed holes. \new{See Fig.~\ref{fig:perc:polygon_examples} for an illustrative example of such a segmented region and the local convex approximations it permits.} Finally, if the particular region allows, we shrink the boundary inwards (and holes outwards) to provide a safety margin. If the inscribed area is not large enough, the plane boundary is accepted without margin. In this way we obtain a margin where there is enough space to do so, but at the same time we do not reject small stepping stones, which might be crucial in certain scenarios.

\subsection{Signed Distance Field}\label{sect:perc:signed_distance_field}
Before computing the SDF, we take advantage of the classification between terrain that will be potentially stepped on and terrain that will not be stepped on. To all cells that are non-steppable, we add a vertical margin of \SI{2}{\centi\meter}, and dilate the elevation by one cell. The latter effectively horizontally inflates all non-steppable areas by the map resolution. This procedure corrects for the problem that edges tend to be underestimated in the provided elevation map.

We use a dense 3D voxel grid, where each voxel contains the value and 3D gradient. The previous motion plan is used to determine the 3D volume where distance information is needed. This volume is a bounding box that contains all collision bodies of the last available plan with a margin of \SI{25}{\centi\meter}. This way, the size and shape of the SDF grid dynamically scales with the motion that is executed. Storing both value and derivative as proposed in \cite{pankert2020perceptive} allows for efficient interpolation during optimization. However, in contrast to \cite{pankert2020perceptive}, where values and gradients are cached after the first call, we opt to precompute the full voxel grid to reduce the computation time during optimization as much as possible. 

This is possible by taking advantage of the extra structure that the 2.5D representation provides. A detailed description of how the SDF can be efficiently computed from an elevation map is given in Appendix~\ref{sect:perc:sdf_appendix}.

\subsection{Torso reference map}\label{sect:perc:torso_reference_layer}
With user input defined as horizontal velocity and an angular rate along the z-direction, it is the responsibility of the controller to decide on the height and orientation of the torso. We would like the torso pose to be positioned in such a way that suitable footholds are in reach for all of the feet. We therefore create a layer that is a smooth interpolation of all steppable regions as described in \cite{jenelten2021TAMOLS}. The use of this layer to generate a torso height and orientation reference is presented in section~\ref{sect:perc:reference_generation}.

\section{Motion planning}
\label{sect:perc:motion_optimization}
In this section, we describe the nonlinear MPC formulation. In particular, we set out to define all components in the following nonlinear optimal control problem:
\begin{subequations}
\begin{align}
     &  \underset{\vu(\cdot)}{\text{minimize}} && \Phi(\vx(T)) + \int_{0}^{T} L(\vx(t),\vu(t), t) \text{ d}t, 
 \label{eq:perc:mpc_cost} \\
    &\text{subject to:} && \vx(0) = \new{\hat{\vx}},  \label{eq:perc:mpc_initial} \\
    & & &  \dot{\vx} =  \vf^c(\vx, \vu, t),  \label{eq:perc:mpc_dynamics} \\
    & & & \vg(\vx,\vu, t) = \zero,  \label{eq:perc:mpc_eqconstraint}
\end{align} \label{eq:perc:mpc_formulation}%
\end{subequations}
where $\vx(t)$ and $\vu(t)$ are the state and the input at time $t$, \new{and $\hat{\vx}$ is the current measured state}. The term $L(\cdot)$ is a time-varying running cost, and $\Phi(\cdot)$ is the cost at the terminal state $\vx(T)$.
The goal is to find a control signal that minimizes this cost subject to the initial condition, $\vx_0$, system dynamics, $\vf^c(\cdot)$, and equality constraints, $\vg(\cdot)$. Inequality constraints are all handled through penalty functions and will be defined as part of the cost function in section~\ref{sect:perc:cost_definition}.

\begin{figure}[!t]
\centering
\includegraphics[width=\columnwidth]{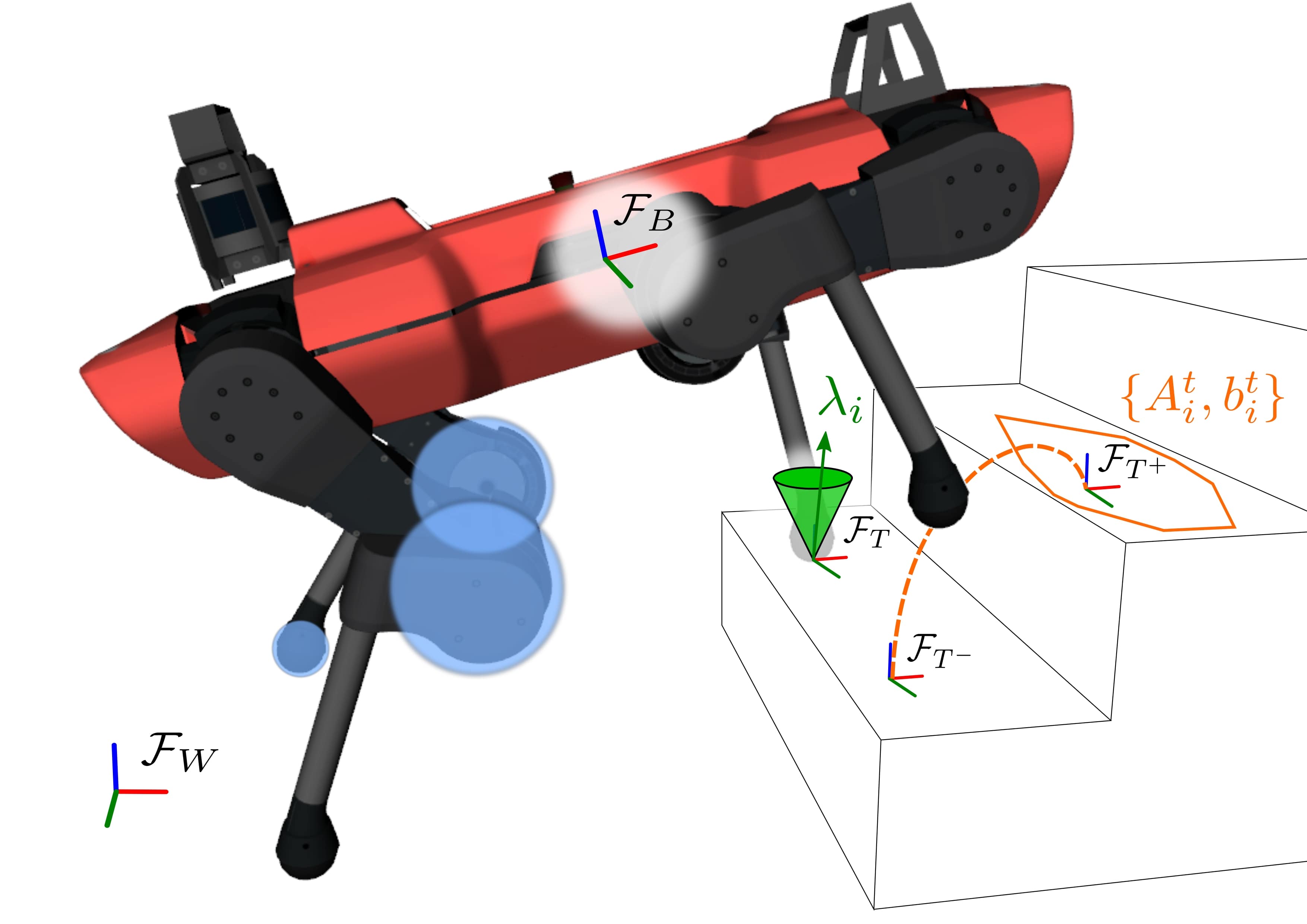}
\caption{Overview of the coordinates frames and constraints used in the definition of the MPC problem. On the front left foot, a friction cone is shown, defined in the terrain frame $\mathcal{F}_T$. On the right front foot, a swing reference trajectory is drawn between the liftoff frame $\mathcal{F}_{T^-}$ and touchdown frame $\mathcal{F}_{T^+}$. Foot placement constraints are defined as a set of half-spaces in the touchdown frame. Stance legs have collision bodies at the knee, as illustrated on the right hind leg, while swing legs have collision bodies on both the foot and the knee, as shown on the left hind leg.}
\label{fig:perc:state_definition}
\end{figure}

\subsection{Robot definition}
We define the generalized coordinates and velocities as:
\begin{equation}
\vq = \left[\vth_{B}^\top, \vp_{B}^\top, \vq_{j}^\top \right]^\top{\mkern-16mu,} \quad \dot{\vq} = \left[\vom_{B}^\top, \vv_{B}^\top, \dot{\vq}_{j}^\top \right]^\top{\mkern-16mu,}
\end{equation}
where $\vth_B\in\R^{3}$ is the orientation of the base frame, $\mathcal{F}_B$, in Euler angles, $\vp_{B}\in\R^3$ is the position of the base in the world frame, $\mathcal{F}_W$. $\vom_{B}\in\R^3$ and $\vv_{B}\in\R^3$ are the angular rate and linear velocity of the base in the body frame $\mathcal{F}_B$. Joint positions and velocities are given by $\vq_j\in\R^{12}$ and $\dot{\vq}_{j}\in\R^{12}$. The collection of all contact forces is denoted by ${\vlambda}\in\R^{12}$. When referring to these quantities per leg, we will use a subscript $i$, e.g. $\vq_i\in\R^{3}$ or ${\vlambda}_i\in\R^{3}$. All subscripts for legs in contact are contained in the set $\mathcal{C}$. A graphical illustration of the robot together with the defined coordinate frames is provided in Fig.~\ref{fig:perc:state_definition}.

\subsection{Torso dynamics}
\label{sect:perc:torso_dynamics}
To derive the torso dynamics used in this work, consider the full rigid body dynamics of the robot,
\begin{equation}
    \vM(\vq)\ddot{\vq}+\vn(\vq,\dot{\vq}) = \vS^\top \vtau + \vtau^{\text{dist}} + \sum_{i\in \mathcal{C}} \vJ^\top_i(\vq) \vlambda_i,
    \label{eq:perc:full_rigid_body_dynamics}
\end{equation}
with inertia matrix $\vM:\R^{18}\to\R^{18\times 18}$, generalized accelerations $\ddot{\vq}\in\R^{18}$, and nonlinear terms $\vn:\R^{18}\times\R^{18}\to\R^{18}$ on the left hand side. The right hand contains the selection matrix $\vS = \left[ \zero_{12\times6}, \,  \vI_{12\times12}  \right] \in\R^{12\times18}$, actuation torques $\vtau\in\R^{12}$,
disturbance forces $\vtau^{\text{dist}}\in\R^{18}$, contact Jacobians $\vJ_i:\R^{18}\to\R^{3\times 18}$, and contact forces ${\vlambda}_i\in\R^{3}$.

For these equations of motion, it is well known that for an articulated system, the underactuated, top 6 rows are of main interest for motion planning \cite{ponton2018ontime}. These so-called centroidal dynamics govern the range of motion that can be achieved \cite{wieber2006holonomy,orin2013centroidal}. Solving the centroidal dynamics for base acceleration gives:
\begin{align}
\begin{bmatrix} \dot{\vom}_{B} \\ \dot{\vv}_{B} \end{bmatrix}
&= 
\vM_{B}^{-1}\left(\vtau^{\text{dist}}_{B} - \vM_{Bj}\Ddot{\vq}_{j} - \vn_{B} + \sum_{i\in \mathcal{C}} \vJ^\top_{B,i} \vlambda_i \right), \\ 
&= \vf_{B}(\vq,\dot{\vq}, \Ddot{\vq}_j, \vlambda, \vtau^{\text{dist}}_{B}),
\label{eq:perc:torso_dynamics}
\end{align}
where $\vM_{B} \in \R^{6\times6}$ is the compound inertia tensor at the top left of $\vM(\vq)$, and $\vM_{Bj} \in \R^{6\times12}$  is the top right block that encodes inertial coupling between the legs and base. The other terms with subscript $B$ correspond to the top 6 rows of the same terms in \eqref{eq:perc:full_rigid_body_dynamics}.

To simplify the torso dynamics, we evaluate this function with zero inertial coupling forces from the joints, i.e.\ $ \vM_{Bj}\Ddot{\vq}_j = \zero$. This simplification allows us to consider the legs only on velocity level and removes joint accelerations from the formulation. From here, further simplifications would be possible. Evaluating the function at a nominal joint configuration and zero joint velocity creates a constant inertia matrix and gives rise to the commonly used single rigid body assumption. While this assumption is appropriate on flat terrain, the joints move far away from their nominal configuration in this work, creating a significant shift in mass distribution and center of mass location.

\subsection{Input loopshaping}
\label{sect:perc:loopshaping}
The bandwidth limitations of the series elastic actuators used in ANYmal pose an additional constraint on the set of motions that are feasible on hardware. Instead of trying to accurately model these actuator dynamics, we use a frequency-dependent cost function to penalize high-frequency content in the contact forces and joint velocity signals \cite{grandia2019frequency}. For completeness, we present here the resulting system augmentation in the time domain:
\begin{align}
    \dot{\vs}_{\lambda} &= \vA_{\lambda} \vs_{\lambda} + \vB_{\lambda} \vnu_{\lambda}, &\dot{\vs}_{j} &= \vA_{j} \vs_{j} + \vB_{j} \vnu_{j}, \label{eq:perc:aug_system}\\
     \vlambda &= \vC_{\lambda} \vs_{\lambda} + \vD_{\lambda} \vnu_{\lambda}, &\dot{\vq}_{j} &= \vC_{j} \vs_{j} + \vD_{j} \vnu_{j}, \notag
\end{align}
where $\vs_{\lambda}$ and $\vs_{j}$ are additional states, and $\vnu_{\lambda}$ and $\vnu_{j}$ are auxiliary inputs, associated with contact forces and joint velocities respectively. When the filters ($\vnu_{\lambda} \rightarrow \vlambda$ and $\vnu_{j} \rightarrow \dot{\vq}_{j}$) are low-pass filters, penalizing the auxiliary input is equivalent to penalizing high frequency content in $\vlambda$ and $\dot{\vq}_{j}$. 

An extreme case is obtained when choosing $\vA_{\lambda} = \vD_{\lambda} = \zero$, $\vB_{\lambda} = \vC_{\lambda} = \vI$, in which case the auxiliary input becomes the derivative, $\dot{\vlambda}$. This reduces to the common system augmentation technique that allows penalization of input rates \cite{rawlings2017model}. 

In our case we allow some direct control ($\vD \neq \zero$) and select $\vA_{\lambda} = \vA_{j} = \zero$, $\vB_{\lambda} =  \vB_{j} = \vI$, $\vC_{\lambda} = \frac{100}{4}\vI$, $\vC_{j} = \frac{50}{3}\vI$, $\vD_{\lambda} = \frac{1}{4}\vI$, $\vD_{j} = \frac{1}{3}\vI$. This corresponds to a progressive increase in cost up to a frequency of \SI{100}{\radian\per\second} for $\vlambda$ and up to \SI{50}{\radian\per\second} for $\dot{\vq}_{j}$, where high frequency components have their cost increased by a factor of $4$ and $3$ respectively.

\subsection{System Dynamics}
We are now ready to define the state vector $\vx \in \R^{48}$ and input vector $\vu \in \R^{24}$ used during motion optimization:
\begin{equation}
\vx = \left[\vth_{B}^\top, \vp_{B}^\top, \vom_{B}^\top, \vv_{B}^\top,  \vq_{j}^\top, \vs_{\lambda}^\top, \vs_{j}^\top \right]^\top{\mkern-16mu,} \quad \vu = \left[\vnu_{\lambda}^\top, \vnu_{j}^\top \right]^\top{\mkern-16mu.}
\end{equation}

Putting together the robot dynamics from section~\ref{sect:perc:torso_dynamics} and system augmentation described in~\ref{sect:perc:loopshaping} gives the continuous time MPC model $\dot{\vx} =  \vf^c(\vx, \vu, t)$:

\begin{equation}
  \frac{\text{d}}{\dt} \begin{bmatrix}
  \vth_{B} \\ \vp_{B} \\ \vom_{B} \\ \vv_{B} \\ \vq_{j} \\ \vs_{\lambda} \\ \vs_{j}
  \end{bmatrix} = 
  \begin{bmatrix}
  \vT(\vth_{B}) \vom_{B} \\
  \vR_B(\vth_{B}) \, \vv_{B} \\
  \vspace{-.5em} \\ \vspace{.5em} % hack to make f appear in the middle
  \vf_{B}(\vq,\dot{\vq}, \zero, \vC_{\lambda} \vs_{\lambda} + \vD_{\lambda} \vnu_{\lambda}, \vtau^{\text{dist}}_{B})
  \\
  \vC_{j} \vs_{j} + \vD_{j} \vnu_{j} \\
  \vA_{\lambda} \vs_{\lambda} + \vB_{\lambda} \vnu_{\lambda} \\
  \vA_{j} \vs_{j} + \vB_{j} \vnu_{j} 
  \end{bmatrix},
  \label{eq:perc:mpc_full_dynamics}
\end{equation}
where $\vT(\vth_{B}):\R^{3}\to\R^{3\times 3}$ provides the conversion between angular body rates and Euler angle derivatives, and $\vR_B(\vth_{B}):\R^{3}\to\R^{3\times 3}$ provides the body to world rotation matrix. The disturbance wrench $\vtau^{\text{dist}}_{B}$ is considered a parameter and is assumed constant over the MPC horizon.

\subsection{Reference generation}
\label{sect:perc:reference_generation}
The user commands 2D linear velocities and an angular rate in the horizontal plane, as well as a desired gait pattern. A full motion and contact force reference is generated to encode these user commands and additional motion preferences into the cost function defined in section~\ref{sect:perc:cost_definition}. 
%Moreover, the sequence of convex foothold constraints is to be extracted from the terrain. 
This process is carried out before every MPC iteration.

As a first step, assuming a constant input along the horizon, a 2D base reference position and heading direction are extrapolated in the world frame. At each point in time, the 2D pose is converted to a 2D position for each hip. The smoothened elevation map, i.e.\ the \textit{torso reference} layer shown in Fig~\ref{fig:perc:perception_overview}, is interpolated at the 2D hip location. The interpolated elevation in addition to a desired nominal height, $h_{nom}$, gives a 3D reference position for each hip. A least-squares fit through the four hip positions gives the 6DoF base reference.

The extracted base reference and the desired gait pattern are used to derive nominal foothold locations. Here we use the common heuristic that the nominal foothold is located below the hip, \new{in gravity-aligned direction}, at the middle of the contact phase \cite{raibert1986legged}. Additionally, for the first upcoming foothold, a feedback on the measured velocity is added:
\begin{equation}
    \vp_{i,nom} = \vp_{i,hip,nom} + \sqrt{\frac{h_{nom}}{g}} (\vv_{B,meas} - \vv_{B,com}),
\end{equation}
where $\vp_{i,nom} \in \R^{3}$ is the nominal foothold, $\vp_{i,hip,nom} \in \R^{3}$ is the nominal foothold location directly below the hip, and $g$ is the gravitational constant. $\vv_{B,meas}$ and $\vv_{B,com}$ are measured and commanded base velocity respectively.

With the nominal foothold locations known, the plane segmentation defined in section~\ref{sect:perc:plane_segmentation} is used to adapt the nominal foothold locations to the perceived terrain. Each foothold is projected onto the plane that is closest and within kinematic limits. Concretely, we pick the reference foothold, $\vp_{i,\new{proj}}$, according to:
\begin{equation}
\argmin_{\vp_{i,\new{proj}} \in \Pi(\vp_{i,nom})} \|\vp_{i,nom} - \vp_{i,\new{proj}}\|^2_2 +  w_{kin} f_{kin}(\vp_{i,\new{proj}}),
\end{equation}
where $\Pi(\vp_{i,nom})$ is \new{a set of candidate points. For each segmented plane we take the point within that region that is closest to the nominal foothold as a candidate. The term} $f_{kin}$ is a kinematic penalty with weight $w_{kin}$ that penalizes the point if the leg extension at liftoff or touchdown is beyond a threshold and if the foothold crosses over to the opposite side of the body. Essentially, this is a simplified version of the foothold batch search algorithm presented in \cite{jenelten2020perceptive}, which searches over cells of the map instead of pre-segmented planes.

After computing all projected footholds, heuristic swing trajectories are computed with two quintic splines; from liftoff to apex and apex to touchdown. The spline is constrained by a desired liftoff and touchdown velocity, and an apex location is selected in such a way that the trajectory clears the highest terrain point between the footholds. Inverse kinematics is used to derive joint position references corresponding to the base and feet references. Finally, contact forces references are computed by dividing the total weight of the robot equally among all feet that are in contact. Joint velocity references are set to zero.

\subsection{Cost \& Soft Inequality Constraints}
\label{sect:perc:cost_definition}
The cost function \eqref{eq:perc:mpc_cost} is built out of several components. The running cost $L(\vx, \vu, t)$ can be split into tracking costs $L_{\vep}$, loopshaping costs $L_{\vnu}$, and penalty costs $L_{\mathcal{B}}$:
\begin{equation}
L = L_{\vep} + L_{\vnu} + L_{\mathcal{B}}.
\end{equation}

The motion tracking cost are used to follow the reference trajectory defined in section~\ref{sect:perc:reference_generation}. Tracking error are defined for the base, $\vep_{B}$, and for each foot ,$\vep_{i}$,
\begin{equation}
    \vep_{B} = \begin{bmatrix}
    \text{log}(\vR_{B}\vR^\top_{B,ref})^\vee \\ 
  \vp_{B} - \vp_{B,ref} \\ \vom_{B} - \vom_{B,ref} \\ \vv_{B} - \vv_{B,ref} 
  \end{bmatrix}, \, \vep_{i} = \begin{bmatrix}
  \vq_{i} - \vq_{i,ref}  \\ 
  \dot{\vq}_{i} - \dot{\vq}_{i,ref} \\
  \vp_{i}  - \vp_{i,ref}  \\ 
  \vv_{i} - \vv_{i,ref}  \\ 
  \vlambda_{i} -  \vlambda_{i,ref} \\ 
  \end{bmatrix},
\end{equation}
where $\text{log}(\vR_{B}\vR^\top_{B,ref})^\vee$ is the logarithmic map of the orientation error, represented as a 3D rotation vector, and $\vp_{i}$ and $\vv_{i}$ are the foot position and velocity in world frame. Together with diagonal, positive definite, weight matrices $\vW_{B}$ and $\vW_{i}$, \new{for which the individual elements are listed in Table~\ref{tab:perc:tracking_weights}}, these errors form the following nonlinear least-squares cost:
\begin{equation}
    L_{\vep} = \frac{1}{2}\|\vep_{B}\|^2_{\vW_{B}} + 
    \sum_{i=1}^4 \frac{1}{2}\|\vep_{i}\|^2_{\vW_{i}}.
    \label{eq:perc:motion_cost}
\end{equation}

\begin{table}[tb]
\centering
\caption{\new{Motion tracking weights}}
\label{tab:perc:tracking_weights}
\begin{tabular}{lrr}
Term & Weights                  \\ \hline
$\text{log}(\vR_{B}\vR^\top_{B,ref})^\vee$  & $(100.0, 300.0, 300.0)$  \\
$ \vp_{B} - \vp_{B,ref} $   &    $(1000.0, 1000.0, 1500.0)$  \\
$ \vom_{B} - \vom_{B,ref} $     & $(10.0, 30.0, 30.0)$               \\
$ \vv_{B} - \vv_{B,ref} $     &  $(15.0, 15.0, 30.0)$               \\
$ \vq_{i} - \vq_{i,ref} $     &  $(2.0, 2.0, 1.0)$ \\
$ \dot{\vq}_{i} - \dot{\vq}_{i,ref} $     & $(0.02, 0.02, 0.01)$  \\
$ \vp_{i}  - \vp_{i,ref} $     &  $(30.0, 30.0, 30.0)$               \\
$ \vv_{i} - \vv_{i,ref} $     &  $(15.0, 15.0, 15.0)$               \\
$ \vlambda_{i} -  \vlambda_{i,ref} $     &  $(0.001, 0.001, 0.001)$ 
\end{tabular}
\end{table}

As discussed in section~\ref{sect:perc:loopshaping}, high-frequency content in joint velocities and contact forces are penalized through a cost on the corresponding auxiliary input. This cost is a simple quadratic cost:
\begin{equation}
    L_{\vnu} = \frac{1}{2} \vnu_{\lambda}^\top \vR_{\lambda} \vnu_{\lambda} + \frac{1}{2} \vnu_{j}^\top \vR_{j} \vnu_{j},
    \label{eq:perc:smoothness}
\end{equation}
where $\vR_{\lambda}$ and $\vR_{j}$ are constant, positive semi-definite, weight matrices. To obtain an appropriate scaling and avoid further manual tuning, these matrices are obtained from the quadratic approximation of the motion tracking cost~\eqref{eq:perc:motion_cost}, with respect to $\vlambda$ and $\dot{\vq}_j$ respectively, at the nominal stance configuration of the robot.

All inequality constraints are handled through the penalty cost. In this work, we use relaxed barrier functions \cite{hauser2006barrier,feller2017stabilizing}. This penalty function is defined as a log-barrier on the interior of the feasible space and switches to a quadratic function at a distance $\delta$ from the constraint boundary.
\begin{equation}
    \mathcal{B}(h)= 
\begin{cases}
    - \mu \ln(h) , & h \geq \delta, \\
    \frac{\mu}{2}\left(\left(\frac{h - 2\delta}{\delta}\right)^2 - 1 \right) - \mu\ln(\delta),  & h < \delta.
\end{cases}
\label{eq:perc:relaxed_barrier}
\end{equation}
The penalty is taken element-wise for vector-valued inequality constraints. The sum of all penalties is given as follows:
\begin{equation}
L_{\mathcal{B}} = \sum_{i=1}^{4} \mathcal{B}_j\left(\vh^{j}_{i}\right) + \sum_{i \in \mathcal{C}} \mathcal{B}_t\left(\vh^{t}_{i} \right) + \mathcal{B}_{\lambda}\left(h^{\lambda}_{i} \right) + \sum_{c \in \mathcal{D}} \mathcal{B}_d\left(h^{d}_{c} \right),
\end{equation}
with joint limit constraints $\vh^{j}_{i}$ for all legs, foot placement and friction cones constraints, $\vh^{t}_{i} $ and $h^{\lambda}_{i}$, for legs in contact, and collision avoidance constraints $h^{d}_{c}$ for all bodies in a set $\mathcal{D}$.

The joint limits constraints contain upper $\{ \overline{\vq}_j$, $\overline{\dot{\vq}}_j$ , $\overline{\vtau}\}$ and lower bounds $\{\underline{\vq}_j$, $\underline{\dot{\vq}}_j$, $\underline{\vtau}\}$ for positions, velocities, and torques:
\begin{equation}
\vh_i^{j} = \begin{bmatrix}
\overline{\vq}_j - \vq_j \\
\vq_j - \underline{\vq}_j \\
\overline{\dot{\vq}}_j - \dot{\vq}_j \\
\dot{\vq}_j - \underline{\dot{\vq}}_j \\
\overline{\vtau} - \vtau \\
\vtau - \underline{\vtau}
\end{bmatrix} \geq \zero,
\label{eq:perc:jointLimits}
\end{equation}
where we approximate the joint torques by considering a static equilibrium in each leg, i.e.\ $\vtau_i = \vJ^\top_{j,i} \vlambda_i$.

The foot placement constraint is a set of linear inequality constraints in task space:
\begin{equation}
\vh_i^{t} = \vA_i^t \cdot \vp_{i} + \vb_i^t \geq \zero,
\label{eq:perc:foothold_position_constraint}
\end{equation}
where $\vA_i^t \in \R^{m\times3}$, and $\vb_i^t \in \R^{m}$ define $m$ half-space constraints in 3D. Each half-space is defined as the plane spanned by an edge of the 2D polygon and the surface normal of the touchdown terrain $\mathcal{F}_{T+}$. The polygon is obtained by initializing all $m$ vertices at the reference foothold derived in section ~\ref{sect:perc:reference_generation} and iteratively displacing them outwards. Each vertex is displaced in a round-robin fashion until it reaches the \new{boundary} of the segmented region or until further movement would cause the polygon to become non-convex. 
%The algorithm terminates once no more vertices can be moved.
Similar to \cite{deits2015computing}, we have favoured the low computational complexity of an iterative scheme over an exact approach of obtaining a convex inner approximation. The \new{first set of} extracted constraints remain unaltered for a foot that is in the second half of the swing phase to prevent last-minute jumps in constraints. 

The friction cone constraint is implemented as:
\begin{equation}
  h^{\lambda}_{i} = \mu_c F_z - \sqrt{F_x^2 + F_y^2 + \epsilon^2} \geq 0, 
    \label{eq:perc:cone}
\end{equation}
with $[F_x, F_y, F_z]^\top = \vR_T^\top \vR_B \vlambda_i$, defining the forces in the local terrain frame. $\mu_c$ is the friction coefficient, and $\epsilon > 0$ is a parameter that ensures a continuous derivative at $\vlambda_i = \zero$, and at the same time creates a safety margin~\cite{grandia2019feedback}.

The collision avoidance constraint is given by evaluation of the SDF at the center of a collision sphere, $\vp_c$, together with the required distance given by the radius, $r_c$, and a shaping function $d_\text{min}(t)$.
\begin{equation}
h^{d}_{c} = d^{SDF}(\vp_c) - r_c - d_\text{min}(t) \geq 0. \label{eq:perc:sdf_inequality}
\end{equation}
The primary use of the shaping function is to relax the constraint if a foot starts a swing phase from below the map. To avoid the robot using maximum velocity to escape the collision, we provide smooth guidance back to free space with a cubic spline trajectory. This happens when the perceived terrain is higher than the actual terrain, for example in case of a soft terrain like vegetation and snow, or simply because of drift and errors in the estimated map. The collision set $\mathcal{D}$ contains collision bodies for all knees and for all feet that are in swing phase, as visualized on the hind legs in Fig.~\ref{fig:perc:state_definition}.

Finally, we use a quadratic cost as the terminal cost in \eqref{eq:perc:mpc_cost}. To approximate the infinite horizon cost incurred after the finite horizon length, we solve a Linear Quadratic Regulator (LQR) problem for the linear approximation of the MPC model and quadratic approximation of the intermediate costs around the nominal stance configuration of the robot. The Riccati matrix $\vS_{\text{LQR}}$ of the cost-to-go is used to define the quadratic cost around the reference state:
\begin{equation}
\Phi(\vx) = \frac{1}{2} \left(\vx - \vx_{ref}(T) \right)^\top \vS_{\text{LQR}} \left(\vx - \vx_{ref}(T)\right).
\end{equation}

\subsection{Equality constraints}
\label{sect:perc:equality_constraints}
For each foot in swing phase, the contact forces are required to be zero: 
\begin{equation}
\vlambda_{i} = \mathbf{0}, \qquad \forall i \notin \mathcal{C}. 
\end{equation}

Additionally, for each foot in contact, the end-effector velocity is constrained to be zero. For swing phases, the reference trajectory is enforced only in the normal direction. This ensures that the foot lifts off and touches down with a specified velocity while leaving complete freedom of foot placement in the tangential direction.
\begin{align*}
&\left\{ 
\begin{array}{ll}
		\vv_{i} = \zero, \quad &\text{if $i \in \mathcal{C}$}, \\
		\vn^\top(t) \left(\vv_{i} - \vv_{i,ref} + k_p (\vp_{i} - \vp_{i,ref})\right) = 0,&\text{if $i \notin \mathcal{C}$},
\end{array}
\right.
\end{align*}
The surface normal, $\vn(t)$, is interpolated over time since liftoff and touchdown terrain can have a different orientation.

\section{Numerical Optimization}
\label{sect:perc:numerical_optimization}

\begin{algorithm}[tb]
\caption{Real-time iteration Multiple-shooting MPC}
\label{alg:SQP}
\footnotesize
\begin{algorithmic}[1]
\State \textbf{Given: } previous solution $\vw_i$
\State Discretize the continous problem to the form of \eqref{eq:perc:NMPC} \label{alg:line:dt}
\State Compute the linear quadratic approximation \eqref{eq:perc:SQP_QPSubproblem} \label{alg:line:LQ}
\State Compute the equality constraint projection \eqref{eq:perc:eq_projection} \label{alg:line:proj}
\State $\delta\tilde{\vw} \gets $ Solve the projected QP subproblem  \eqref{eq:perc:SQP_ProjectedQPSubproblem} \label{alg:line:hpipmsolve}
\State $\delta\vw \gets \vP \delta\tilde{\vw} + \vp$, back substitution using \eqref{eq:perc:qp_subspace} \label{alg:line:backproj}
\State $\vw_{i+1} \gets $ Line-Search($\vw_i$, $\delta \vw$), (Algorithm~\ref{alg:ls}) \label{alg:line:ls}
\end{algorithmic}
\end{algorithm}

We consider a direct multiple-shooting approach to transforming the continuous optimal control problem into a finite-dimensional nonlinear program (NLP) \cite{bock1984multiple}. Since MPC computes control inputs over a receding horizon, successive instances of \eqref{eq:perc:NMPC} are similar and can be efficiently warm-started when taking an SQP approach \new{by shifting the previous solution. For new parts of the shifted horizon, for which no initial guess exists, we repeat the final state of the previous solution and initialize the inputs with the references generated in section~\ref{sect:perc:reference_generation}}. Additionally, we follow the real-time iteration scheme where only one SQP step is performed per MPC update~\cite{diehl2002realtime}. In this way, the solution is improved across consecutive instances of the problem, rather than iterating until convergence for each problem. 

As an overview of the approach described in the following sections, a pseudo-code is provided in Algorithm~\ref{alg:SQP}, referring to the relevant equations used at each step. Except for the solution of the QP in line~\ref{alg:line:hpipmsolve}, all steps of the algorithm are parallelized across the shooting intervals. The QP is solved using HPIPM~\cite{frison2020hpipm}.

\subsection{Discretization}
The continuous control signal $\vu(t)$ is parameterized over subintervals of the prediction horizon $[t,t + T]$ to obtain a finite-dimensional decision problem. This creates a grid of nodes $k \in \{0, \shortdots, N\}$ defining control times $t_k$ separated by intervals of duration $\delta t \approx T/(N-1)$. Around gait transitions, $\delta t$ is slightly shortened or extended such that a node is exactly at the gait transition. 

In this work, we consider a piecewise constant, or zero-order-hold, parameterization of the input. Denoting $\vx_k = \vx(t_k)$ and integrating the continuous dynamics in \eqref{eq:perc:mpc_full_dynamics} over an interval leads to a discrete time representation of the dynamics:
\begin{equation}
   \vf_k^d(\vx_k, \vu_k) = \vx_k + \int_{t_k}^{t_k+\delta t} \vf^c(\vx(\tau), \vu_k, t) \text{ d}\tau.
   \label{eq:perc:discrete-dynamics}
\end{equation}
The integral in \eqref{eq:perc:discrete-dynamics} is numerically approximated with an integration method of choice to achieve the desired approximation accuracy of the evolution of the continuous time system under the zero-order-hold commands. We use an explicit second-order Runge-Kutta scheme.

% Hard constrained NMPC (without slack variables):
% \begin{subequations}
% \begin{align}
% % cost
% 	\underset{\begin{subarray}{c}
% 		X, U
% 		\end{subarray}}{\min} &&
% 	 l_N(x_N) + \sum_{k=0}^{N-1}& l_k(x_k, u_k)  \\
% 	% initial condition
% 	\text{s.t} && x_0 - \hat{x} &= 0, \\
% 	%% dynamics
% 	&& x_{k+1} - f^d_k(x_k, u_k ) &= 0, \quad k = 0,\dots,\,N-1,\\
% 	% constraints
% 	&& h_k(x_k, u_k) &\leq 0, \quad k = 0,\dots,\,N-1, \\
% 	&& h_N(x_N) &\leq 0, 
% \end{align}
% \end{subequations}

The general nonlinear MPC problem presented below can be formulated by defining and evaluating a cost function and constraints on the grid of nodes. 
\begin{subequations}
\label{eq:perc:NMPC} %
\begin{flalign}
	% initial condition
	&\underset{\begin{subarray}{c}
		\vX, \vU
		\end{subarray}}{\min}& \mathclap{\,\, \Phi(\vx_N)  + \sum_{k=0}^{N-1} l_k(\vx_k, \vu_k) ,} \\
	\label{eq:perc:NMPC-ic}
	&\quad\text{s.t.}&  \vx_0 - \hat{\vx} &= \zero, \\
	%% dynamics
	\label{eq:perc:NMPC-dyn}
	&& \vx_{k+1} - \vf^d_k(\vx_k, \vu_k) &= \zero, &&k = 0,\shortdots,N\!-\!1,\\
	% constraints
	&& \vg_k(\vx_k, \vu_k) & = \zero, &&k = 0,\shortdots,N\!-\!1,
\end{flalign} 
\end{subequations}
where $\vX = [\vx_0^\top, \dots \vx_N^\top]^\top$, and $\vU = [\vu_0^\top, \dots \vu_{N-1}^\top]^\top$, are the sequences of state and input variables respectively. The nonlinear cost and constraint functions $l_k$, and $\vg_k$, are discrete sample of the continuous counterpart. Collecting all decision variables into a vector, $\vw = [\vX^\top, \vU^\top]^\top$, problem \eqref{eq:perc:NMPC} can be written as a general NLP:
\begin{equation}
    \underset{\vw}{\min} \quad \phi(\vw), \quad  \text{s.t.} \quad
        \begin{bmatrix}
        \vF(\vw) \\
        \vG(\vw)
        \end{bmatrix}  = \zero, 
\label{eq:perc:NLP}
\end{equation}
where $\phi(\vw)$ is the cost function, $\vF(\vw)$ is the collection of initial state and dynamics constraints, and $\vG(\vw)$ is the collection of all general equality constraints.

\subsection{Sequential Quadratic Programming (SQP)}
SQP based methods apply Newton-type iterations to Karush-Kuhn-Tucker (KKT) optimality conditions, assuming some regularity conditions on the constraints \cite{mangasarian1967fritz}. The Lagrangian of the NLP in \eqref{eq:perc:NLP} is defined as:
\begin{equation}
    \mathcal{L}(\vw, \vlambda_\vG, \vlambda_\vH) = \phi(\vw) + \vlambda_\vF^\top \vF(\vw) + \vlambda_\vG^\top \vG(\vw),
    \label{eq:perc:NLP_lagrangian}
\end{equation}
with Lagrange multipliers $\vlambda_\vF$ and $\vlambda_\vG$, corresponding to the dynamics and equality constraints. The Newton iterations can be equivalently computed by solving the following potentially non-convex QP \cite{nocedal2006numerical}:
\begin{subequations}
\label{eq:perc:SQP_QPSubproblem}
\begin{align}
% cost
	\underset{\begin{subarray}{c}
		\delta \vw
		\end{subarray}}{\min} & \quad
	 \nabla_\vw \phi(\vw_i)^\top \delta \vw + \frac{1}{2} \delta \vw^\top \vB_i \delta \vw,  \label{eq:perc:SQP-qp-cost} & \\
% Constraints	 
	\quad\text{s.t} & \quad  \vF(\vw_i) + \nabla_\vw \vF(\vw_i)^\top \delta \vw = \zero, & \label{eq:perc:SQP-qp-dynconstr} \\
	& \quad  \vG(\vw_i) + \nabla_\vw \vG(\vw_i)^\top \delta \vw = \zero, & \label{eq:perc:SQP-qp-eqconstr} 
\end{align}
\end{subequations}
where the decision variables, $\delta \vw = \vw - \vw_i$, define the update step relative to the current iteration $\vw_i$, and the Hessian $\vB_i = \nabla^2_\vw\mathcal{L}(\vw_i,\vlambda_\vF, \vlambda_\vG)$. Computing the solution to \eqref{eq:perc:SQP_QPSubproblem} provides a candidate decision variable update, $\delta \vw_i$, and updated Lagrange multipliers. 

\subsection{Quadratic Approximation Strategy}
As we seek to deploy MPC on dynamic robotic platforms, it is critical that the optimization problem in \eqref{eq:perc:SQP_QPSubproblem} is well conditioned and does not provide difficulty to numerical solvers. In particular, when $\vB_i$ in \eqref{eq:perc:SQP-qp-cost} is positive semi-definite (p.s.d), the resulting QP is convex and can be efficiently solved \cite{kouzoupis2018recent}.

To ensure this, an approximate, p.s.d Hessian is used instead of the full Hessian of the Lagrangian. For the tracking costs \eqref{eq:perc:motion_cost}, the objective function has a least-squares form in which case the Generalized Gauss-Newton approximation,
\begin{equation}
  \nabla_\vw^2 \left( \frac{1}{2}\|\vep_i(\vw)\|^2_{\vW_i} \right)  \approx \nabla_\vw \vep_i(\vw)^\top \vW_i \nabla_\vw \vep_i(\vw),  
  \label{eq:perc:GaussNewton}
\end{equation}
proves effective in practice \cite{houska2011auto}. Similarly, for the soft constraints, we exploit to convexity of the penalty function applied to the nonlinear constraint \cite{verschueren2016exploiting}:
\begin{equation}
  \nabla_\vw^2 \left( \mathcal{B}(\vh(\vw)) \right)  \approx \nabla_\vw \vh(\vw)^\top \nabla_\vh^2 \mathcal{B}(\vh(\vw)) \nabla_\vw \vh(\vw),  
  \label{eq:perc:SCQP}
\end{equation}
where the diagonal matrix $\nabla_\vh^2 \mathcal{B}(\vh(\vw))$ maintains the curvature information of the convex penalty functions. The contribution of the constraints to the Lagrangian in~\eqref{eq:perc:NLP_lagrangian} is ignored in the approximate Hessian since we do not have additional structure that allows a convex approximation.

\subsection{Constraint Projection}
The equality constraints in \ref{sect:perc:equality_constraints} were carefully chosen to have full row rank w.r.t. the control inputs, such that, after linearization, $\nabla_\vw \vG(\vw_i)^\top$ has full row rank in \eqref{eq:perc:SQP-qp-eqconstr}. This means that the equality constraints can be eliminated before solving the QP through a change of variables~\cite{nocedal2006numerical}:
\begin{equation}
    \delta\vw = \vP \delta\tilde{\vw} + \vp, \label{eq:perc:qp_subspace} 
\end{equation}
where the linear transformation satisfies
\begin{equation}
    \nabla_\vw \vG(\vw_i)^\top\vP = \zero, \quad  \nabla_\vw \vG(\vw_i)^\top \vp = -\vG(\vw_i).
    \label{eq:perc:eq_projection}
\end{equation}
After substituting \eqref{eq:perc:qp_subspace} into \eqref{eq:perc:SQP_QPSubproblem}, the following QP is solved w.r.t. $\delta\tilde{\vw}$. 
\begin{subequations}
\label{eq:perc:SQP_ProjectedQPSubproblem}
\begin{align}
% cost
	\underset{\begin{subarray}{c}
		\delta \tilde{\vw}
		\end{subarray}}{\min} & \quad
	 \nabla_{\tilde{\vw}} \tilde{\phi}(\vw_i)^\top \delta \tilde{\vw} + \frac{1}{2} \delta \tilde{\vw}^\top \tilde{\vB}_i \delta \tilde{\vw},  \label{eq:perc:SQP-pr-qp-cost} & \\
% Constraints	 
	\quad\text{s.t} & \quad  \tilde{\vF}(\vw_i) + \nabla_{\tilde{\vw}} \tilde{\vF}(\vw_i)^\top \delta \tilde{\vw} = \zero. & \label{eq:perc:SQP-pr-qp-dynconstr} 
\end{align}
\end{subequations}

Because each constraint applies only to the variables at one node $k$, the coordinate transformation  maintains the sparsity pattern of an optimal control problem and can be computed in parallel. Since this projected problem now only contains costs and system dynamics, solving the QP only requires one Ricatti-based iteration\cite{frison2020hpipm}. The full update $ \delta\vw$ is then obtained through back substitution into \eqref{eq:perc:qp_subspace}.

\subsection{Line-Search}
To select an appropriate stepsize, we employ a line-search based on the filter line-search used in IPOPT \cite{wachter2006implementation}. In contrast to a line-search based on a merit function, where cost and constraints are combined to one metric, the main idea is to ensure that each update either improves the constraint satisfaction or the cost function. The constraint satisfaction $\theta(\vw)$ is measured by taking the norm of all constraints scaled by the time discretization:
\begin{equation}
    \theta(\vw) = \delta t \left|\left| \begin{bmatrix} 
        \vF(\vw)^\top, 
        \vG(\vw)^\top
        \end{bmatrix}^\top\right|\right|_2.
        \label{eq:perc:constraint_metric}
\end{equation}

In case of high or low constraint satisfaction, the behavior is adapted: When the constraint is violated beyond a set threshold, $\theta_\text{max}$, the focus changes purely to decreasing the constraints; when constraint violation is below a minimum threshold, $\theta_\text{min}$, the focus changes to minimizing costs. 

Compared to the algorithm presented in \cite{wachter2006implementation}, we remove recovery strategies and second-order correction steps, for which there is no time in the online setting. Furthermore, the history of iterates plays no role since we perform only one iteration per problem.

The simplified line-search as used in this work is given in Algorithm~\ref{alg:ls} and contains three distinct branches in which a step can be accepted. The behavior at high constraint violation is given by line~\ref{ls:line:max}, where a step is rejected if the new constraint violation is above the threshold and worse than the current violation. The switch to the low constraint behavior is made in line~\ref{ls:line:min}: if both new and old constraint violations are low and the current step is in a descent direction, we require that the cost decrease satisfies the Armijo condition in line~\ref{ls:line:armijo}. Finally, the primary acceptance condition is given in line~\ref{ls:line:dual}, where either a cost or constraint decrease is requested. The small constants $\gamma_\phi$, and $\gamma_\theta$ are used to fine-tune this condition with a required non-zero decrease in either quantity.

\begin{algorithm}[!t]
\footnotesize
	\caption{Backtracking Line-Search}
	\label{alg:ls}
	\begin{algorithmic}[1]
	    \State \textbf{Hyperparameters:} $
	    \alpha_\text{min}=10^{-4},
	    \theta_\text{max}=10^{-2}, 
	    \theta_\text{min}=10^{-6},
	    \eta=10^{-4},
	    \gamma_\phi=10^{-6}, 
	    \gamma_\theta=10^{-6}, 
	    \gamma_\alpha=0.5
	    $
		\State $\alpha\leftarrow 1.0$
		\State $\theta_{k} \gets \theta(\vw_i)$
		\State $\phi_{k} \gets \phi(\vw_i)$
		\State Accepted $\gets$ False
		\While{\textit{Not} Accepted and $\alpha\geq \alpha_{\text{min}}$}
		\State $\theta_{i+1} \gets \theta(\vw_i + \alpha\delta\vw)$
		\State $\phi_{i+1} \gets \phi(\vw_i + \alpha\delta\vw)$
		\If {$\theta_{i+1} > \theta_{\text{max}}$ \label{ls:line:max}} 
				\If {$\theta_{i+1} < (1 - \gamma_\theta) \theta_i$ \label{ls:line:constraint}} 
		            \State Accepted $\gets$ True
		        \EndIf
		\ElsIf {$\text{max}(\theta_{i+1},\theta_i) <  \theta_{\text{min}}$ and $\nabla \phi(\vw_i)^\top \delta\vw < 0 $ \label{ls:line:min} }   
		        \If {$\phi_{i+1} < \phi_i + \eta \alpha\nabla \phi(\vw_i)^\top \delta\vw$} \label{ls:line:armijo}
		            \State Accepted $\gets$ True
		        \EndIf        
		\Else {}
		        \If {$\phi_{i+1} < \phi_i - \gamma_\phi \theta_i$ or $
		        \theta_{i+1} < (1 - \gamma_\theta) \theta_i
		        $ \label{ls:line:dual}} 
		            \State Accepted $\gets$ True
		        \EndIf
		\EndIf    
		
		\If {\textit{Not} Accepted}
		    \State $\alpha\leftarrow \gamma_\alpha \alpha$
		\EndIf
		\EndWhile
		\If {Accepted}
		    \State $\vw_{i+1} \gets \vw_i + \alpha\delta\vw$
		\Else
		    \State $\vw_{i+1} \gets \vw_i$
		\EndIf
	\end{algorithmic} 
\end{algorithm}

\section{Motion Execution}
\label{sect:perc:motion_execution}
The optimized motion planned by the MPC layer consists of contact forces and desired joint velocities. We linearly interpolate the MPC motion plan at the \SI{400}{\hertz} execution rate and apply the feedback gains derived from the Riccati backward pass to the measured state~\cite{grandia2019feedback}. The corresponding torso acceleration is obtained through \eqref{eq:perc:torso_dynamics}. The numerical derivative of the planned joint velocities is used to determine a feedforward joint acceleration. A high-frequency whole-body controller (WBC) is used to convert the desired acceleration tasks into torque commands~\cite{sentis2006wholebody,saab2013dynamic,bellicoso2016perception}. A generalized momentum observer is used to estimate the contact state~\cite{bledt2018contact}. Additionally, the estimated external torques are filtered and added to the MPC and WBC dynamics as described in~\cite{jenelten2021TAMOLS}. \new{We use the same filter setup as shown in Fig. 13. of \cite{jenelten2021TAMOLS}.}

\subsection{Event based execution}
Inevitably, the measured contact state will be different from the planned contact state used during the MPC optimization. In this case, the designed contact forces cannot be provided by the whole-body controller. We have implemented simple reactive behaviors to respond to this situation and provide feedback to the MPC layer. 

In case there is a planned contact, but no contact is measured, we follow a downward \textit{regaining} motion for that foot. Under the assumption that the contact mismatch will be short, the MPC will start a new plan again from a closed contact state. Additionally, we propagate the augmented system in \eqref{eq:perc:aug_system} with the information that no contact force was generated, i.e. $\zero \overset{!}{=} \vC_{\lambda} \vs_{\lambda} + \vD_{\lambda} \vnu_{\lambda}$. In this way, the MPC layer will generate contact forces that maintain the requested smoothness w.r.t. the executed contact forces.

When contact is measured, but no contact was planned, the behavior depends on the planned time till contact. If contact was planned to happen soon, the measured contact is sent to the MPC to generate the next plan from that early contact state. \new{In the meantime, the WBC maintains a minimum contact force for that foot.} If no upcoming contact was planned, the measured contact is ignored.

\subsection{Whole-body control}
The whole-body control (WBC) approach considers the full nonlinear rigid body dynamics of the system in \eqref{eq:perc:full_rigid_body_dynamics}, including the estimate of disturbance forces. Each task is formulated as an equality constraint, inequality constraint, or least-squares objective affine in the generalized accelerations, torques, and contact forces. While we have used a hierarchical resolution of tasks in the past~\cite{bellicoso2016perception}, in this work, we instead use a single QP and trade off the tracking tasks with weights. We found that a strict hierarchy results in a dramatic loss of performance in lower priority tasks when inequalities constraints are active. Additionally, the complexity of solving multiple QPs and null-space projections in the hierarchical approach is no longer justified with the high quality motion reference coming from the MPC. 

The complete list of tasks is given in Table~\ref{tab:perc:controllertasks}. The first two blocks of tasks enforce physical consistency and inequality constraints on torques, forces, and joint configurations. The joint limit constraint is derived from an exponential Control Barrier Function (CBF) \cite{nguyen2016exponential} on the joint limits, $\underline{\vq}_j \leq \vq_j \leq \overline{\vq}_j$, resulting in the following joint acceleration constraints:
\begin{align}
\ddot{\vq}_j + (\gamma_1 + \gamma_2) \dot{\vq}_j + \gamma_1 \gamma_2 (\vq_j - \underline{\vq}_j )\geq \zero, \\ 
-\ddot{\vq}_j - (\gamma_1 + \gamma_2) \dot{\vq}_j + \gamma_1 \gamma_2 (\overline{\vq}_j - \vq_j ) \geq \zero,
\end{align}
with scalar parameters $\gamma_1 > 0, \gamma_2 > 0$. These CBF constraints guarantee that the state constraints are satisfied for all time and under the full nonlinear dynamics of the system~\cite{ames2014control}.

For the least-square tasks, we track swing leg motion with higher weight than the torso reference. This prevents that the robot exploits the leg inertia to track torso references in underactuated directions, and it ensures that the foot motion is prioritized over torso tracking when close to kinematics limits. Tracking the contact forces references with a low weight regulates the force distribution in case the contact configuration allows for internal forces. 

Finally, the torque derived from the whole-body controller, $\vtau_{\text{wbc}}\in\R^{12}$, is computed. To compensate for model uncertainty for swing legs, the integral of joint acceleration error with gain $K > 0$ is added to the torque applied to the system: 
\begin{align}
\vtau_i &= \vtau_{i,\text{wbc}} - K \int_{t^{sw}_{0}}^t \left( \ddot{\vq}_i - \ddot{\vq}_{i,\text{wbc}}  \right) \text{d}t, \\
&\new{= \vtau_{i,\text{wbc}} - K \left(\dot{\vq}_i - \dot{\vq}_i(t^{sw}_{0}) -  \int_{t^{sw}_{0}}^t \ddot{\vq}_{i,\text{wbc}} \text{d}t \right),} \label{eq:joint_integral_impl}
\end{align}
where $t^{sw}_{0}$ is the start time of the swing phase. \new{The acceleration integral can be implemented based on the measured velocity $\dot{\vq}_i$ and the velocity at the start of the swing phase, $\dot{\vq}_i(t^{sw})$, as shown in \eqref{eq:joint_integral_impl}. Futhermore, the feedback term is saturated to prevent integrator windup.} For stance legs, a PD term is added around the planned joint configuration and contact consistent joint velocity.

\begin{table}[bt]
\centering
\caption{Whole-body control tasks}
\begin{tabular}{|c|l|}
\hline
Type &Task \\
\hline
 \multirow{2}{*}{$=$} 	& Floating base equations of motion. \\
  						& No motion at the contact points. \\
\hline
 \multirow{3}{*}{$\geq$} 	& Torque limits. \\
							& Friction cone constraint. \\
							& Joint limit barrier constraint. \\
\hline
 \multirow{3}{*}{$w_i^2 \| \cdot \|^2$} 	& Swing leg motion tracking ($w_i = 100.0$). \\
 									& Torso linear and angular acceleration ($w_i = 1.0$). \\
									& Contact force tracking. ($w_i = 0.01$). \\
\hline
\end{tabular}
\label{tab:perc:controllertasks}
\end{table}

\section{Results}
\label{sect:perc:results}

ANYmal is equipped with either two dome shaped Robo-Sense bpearl LiDARs, mounted in the front and back of the torso, or with four Intel RealSense D435 depth cameras mounted on each side of the robot. Elevation mapping runs at \SI{20}{\hertz} on an onboard GPU (Jetson AGX Xavier). Control and state estimation are executed on the main onboard CPU (Intel i7-8850H,2.6 GHz, Hexa-core) at \SI{400}{\hertz}, asynchronously to the MPC optimization which is triggered at \SI{100}{\hertz}. Four cores are used for parallel computation in the MPC optimization. A time horizon of $T =$ \SI{1.0}{\second} is used with a nominal time discretization of $\delta t \approx$ \SI{0.015}{\second}, with a slight variation due to the adaptive discretization around gait transitions. Each multiple-shooting MPC problem therefore contains around $5000$ decision variables. Part (\textbf{A}) and (\textbf{B}) of perception pipeline in Fig.~\ref{fig:perc:perception_overview} are executed on a second onboard CPU of the same kind and provides the precomputed layers over Ethernet.

To study the performance of the proposed controller, we report results in different scenarios and varying levels of detail. \new{All perception, MPC, and WBC parameters remain constant throughout the experiments and are the same for simulation and hardware. An initial guess for these parameters was found in simulation, and we further fine-tuned them on hardware.} First, results for the perception pipeline in isolation are presented in section~\ref{sect:perc:results_perception}. Second, we validate the major design choices in simulation in section~\ref{sect:perc:results_simulation}. Afterward, the proposed controller is put to the test in challenging simulation, as well as hardware experiments in section~\ref{sect:perc:results_hardware}. All experiments are shown in the supplemental video\cite{video}. Finally, known limitations are discussed in section~\ref{sect:perc:results_limitations}. 

\subsection{Perception Pipeline}
\label{sect:perc:results_perception}

\begin{figure}[!t]
\centering
\includegraphics[width=\columnwidth,trim=0 0 0 0,clip]{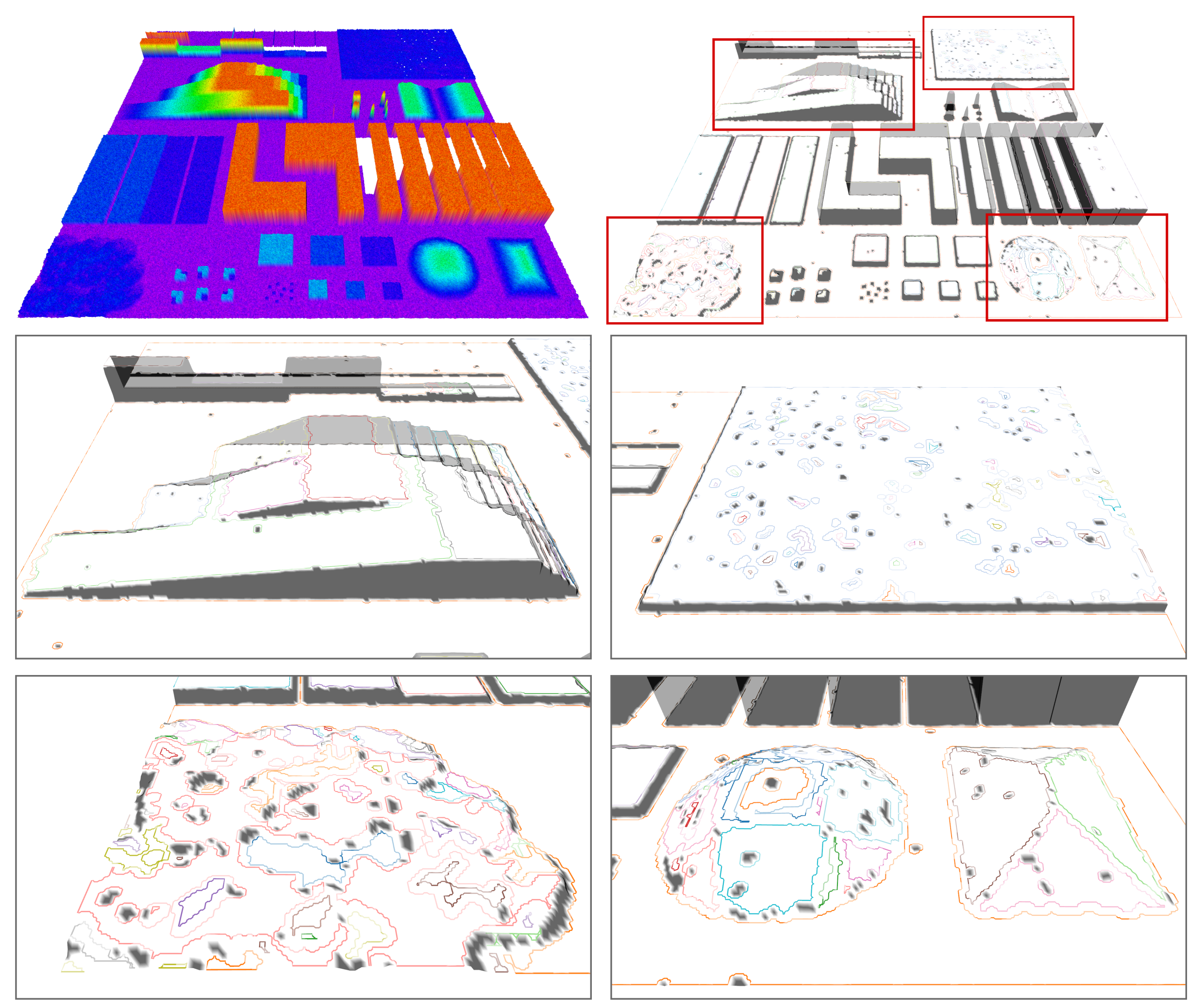}
\caption{Evaluation of the plane segmentation on a demo terrain \cite{fankhauser2016universal}. The shown map has a true size of \SI{20}{}$\times$\SI{20}{}$\times$\SI{1}{\meter} with a resolution of \SI{4}{\centi\meter}. Top left shows the elevation map with additive uniform noise of $\pm$\SI{2}{\centi\meter} plus Gaussian noise with a standard deviation of \SI{2}{\centi\meter}. Top right shows the map after inpainting, filtering, steppability classification, and plane segmentation. Below, four areas of interest are shown. Their original location in the map is marked in the top right image.}
\label{fig:perc:demo_terrain}
\end{figure}

The output of the steppability classification and plane segmentation (part A in Fig.~\ref{fig:perc:perception_overview}) for a demo terrain is shown in Fig.~\ref{fig:perc:demo_terrain}. This terrain is available as part of the gridmap library and contains a collection of slopes, steps, curvatures, rough terrain, and missing data. The left middle image shows that slopes and steps are, in general, well segmented. In the bottom right image, one sees the effect of the plane segmentation on a curved surface. In those cases, the terrain will be segmented into a collection of smaller planes. Finally, the rough terrain sections shown in the right middle and bottom left image show that the method is able to recognize such terrain as one big planar section as long as the roughness is within the specified tolerance. These cases also show the importance of allowing holes in the segmented regions, making it possible to exclude just those small regions where the local slope or roughness is outside the tolerance. A global convex decomposition of the map would result in a much larger amount of regions.

\begin{figure}[!t]
\centering
\includegraphics[width=0.8\columnwidth,trim=0 0 0 0,clip]{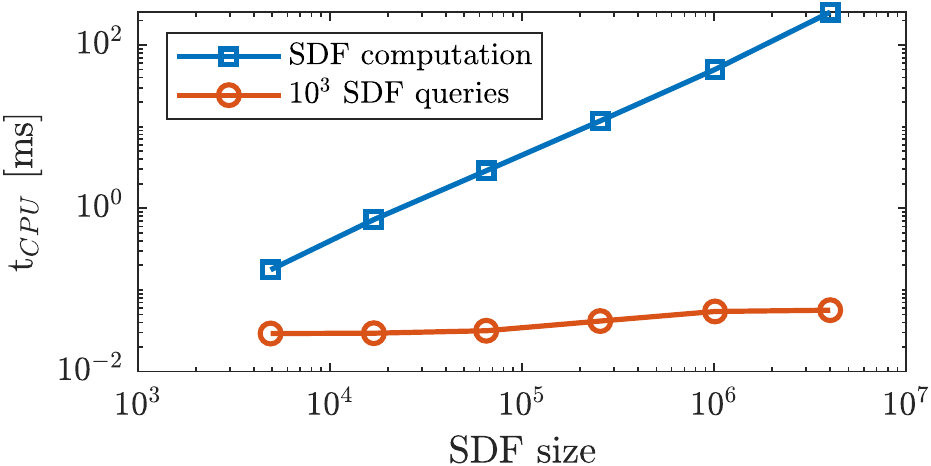}
\caption{Computation time for constructing and querying the signed distance field. Submaps of the terrain in Fig.~\ref{fig:perc:demo_terrain} are used. \textit{SDF size} on the horizontal axis denotes the total amount of data points in the SDF (width$\times$length$\times$height). The query time is reported for the total of $10^3$ random queries for the interpolated value and derivative.}
\label{fig:perc:sdf_computation}
\end{figure}

The computation time for the construction and querying of the signed distance field is benchmarked on sub-maps of varying sizes extracted from the demo map, see Fig.~\ref{fig:perc:sdf_computation}. As expected, the construction time scales linearly with the SDF size, and the query time is constant with a slight increase when the memory size exceeds a cache level. During runtime, the local SDF size is typically below $10^5$ voxels, resulting in a computation time well below \SI{10}{\milli\second}. Together with the map update rate of \SI{20}{\hertz}, the proposed method provides the SDF at an order of magnitude faster than methods that maintain a general 3D voxel grid, with update rated reported around \SI{1}{\hertz}\cite{pankert2020perceptive}. Per MPC iteration, around $10^3$ SDF queries are made, making the SDF query time negligible compared to the total duration of one MPC iteration.

\subsection{Simulation}
\label{sect:perc:results_simulation}

\subsubsection{Collision avoidance}
\begin{figure}[!tb]
    \centering
    \begin{minipage}{.49\columnwidth}
        \centering
        \includegraphics[width=\linewidth,trim=550 300 430 150,clip]{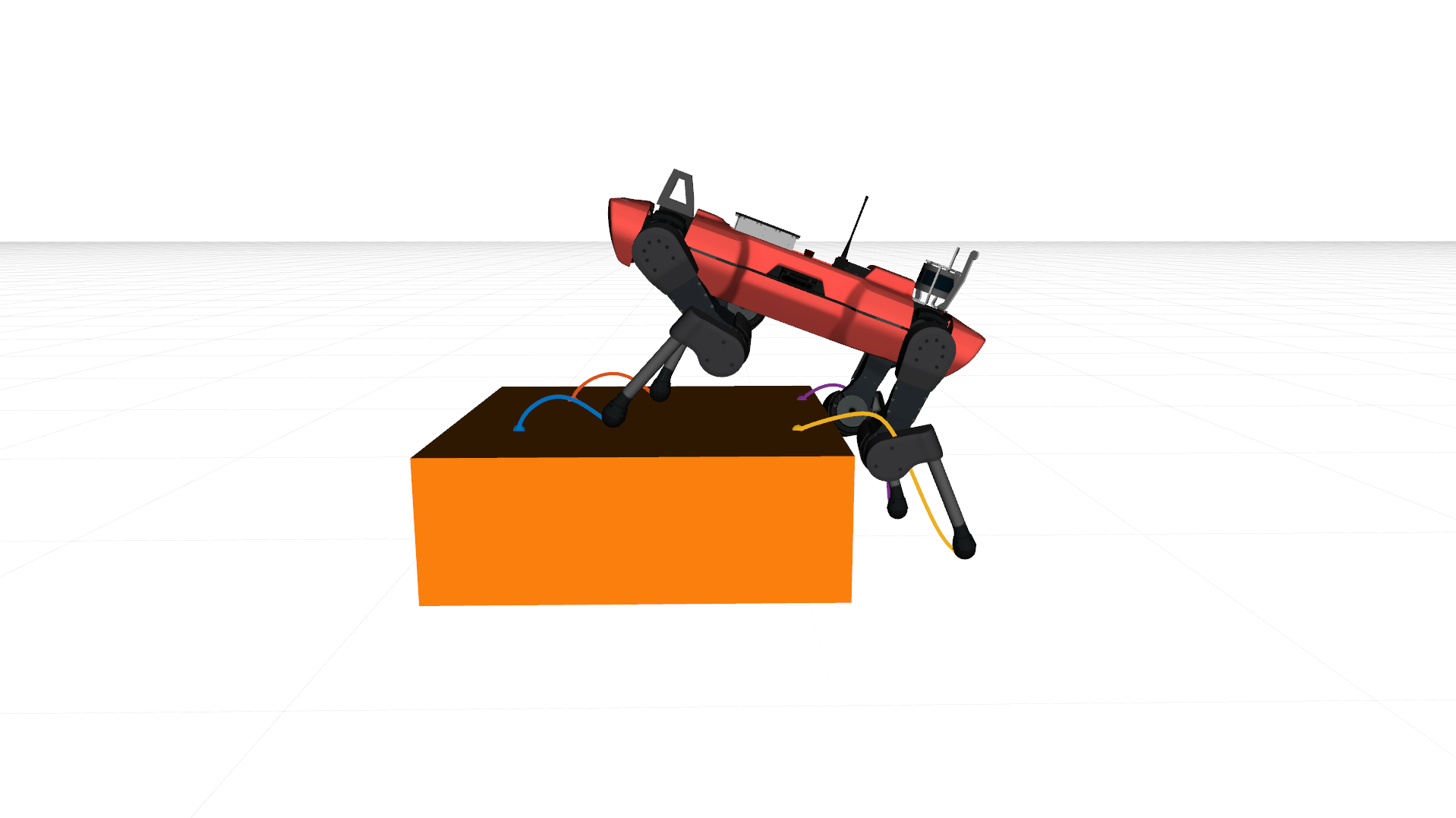}
    \end{minipage}
    \begin{minipage}{.49\columnwidth}
        \centering
        \includegraphics[width=\linewidth,trim=550 300 430 150,clip]{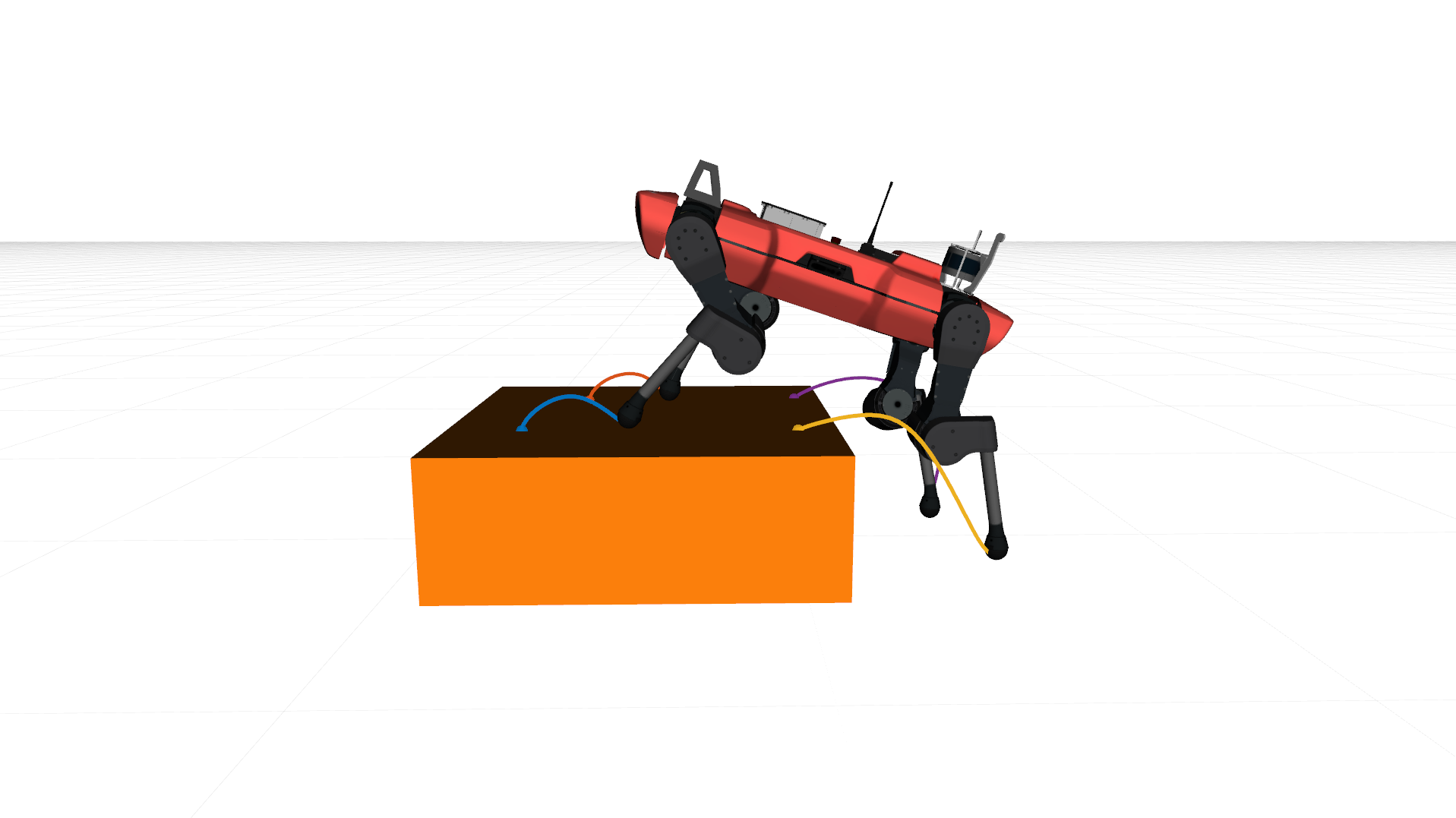}
    \end{minipage}
    \caption{ANYmal stepping up a box of \SI{35}{\centi\meter}. Left: Without considering knee collisions. Right: Knee collision included in the optimization.}
    \label{fig:perc:box_climb_knee_col}
\end{figure}

To highlight the importance of considering knee collisions with the terrain, the robot is commanded to traverse a box of \SI{35}{\centi\meter} with a trotting gait at \SI{0.25}{\meter\per\second}. Fig.~\ref{fig:perc:box_climb_knee_col} compares the simulation result of this scenario with and without the knee collisions considered. The inclusion of knee collision avoidance is required to successfully step up the box with the hind legs. As shown in the figure, the swing trajectories are altered. Furthermore, the base pose and last stepping location before stepping up are adjusted to prepare for the future, showing the benefit of considering all degrees of freedom in one optimization. \new{Similarly, on the way down, the foothold optimization (within  constraints) allows that the feet are placed away from the step, avoiding knee collisions while stepping down.}

\begin{figure}[!tb]
\centering
\includegraphics[width=0.90\columnwidth]{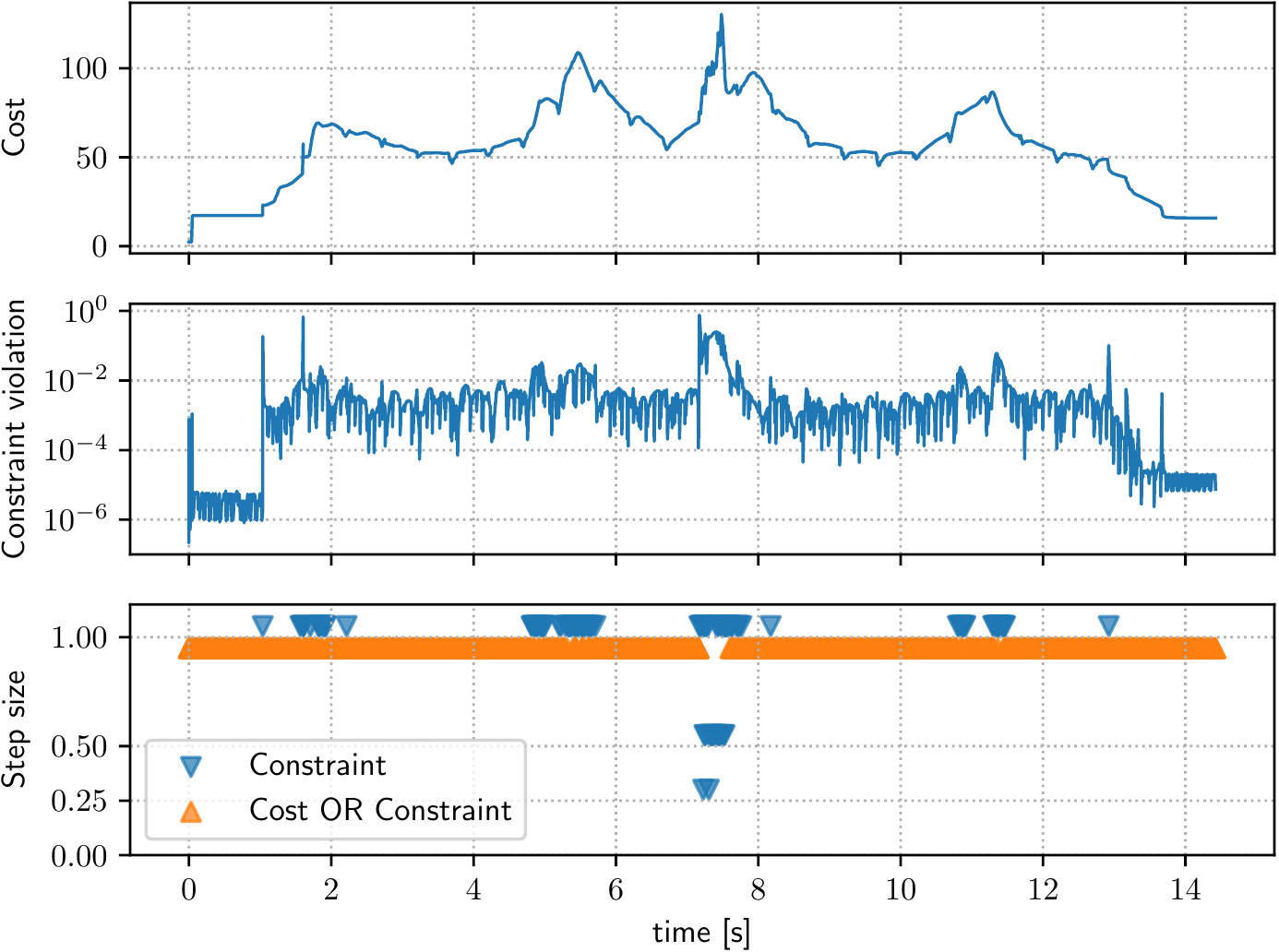}
\caption{Solver status during the box traversal motion (including knee collision avoidance). The first and second plots show the total cost, and constraint violation according to~\eqref{eq:perc:constraint_metric}, after each iteration. The bottom plot shows the stepsize and the line-search branch that led to the step acceptance. `Constraint` refers to a step accepted in the high constraint violation branch in line~\ref{ls:line:max} of Algorithm \ref{alg:ls}, \new{`Cost OR Constraint`} refers to the branch where either cost or constraint decrease is accepted in line~\ref{ls:line:dual}. \new{Note that the low constraint violation branch, line~\ref{ls:line:min}, did not occur in this experiment.}}
\label{fig:perc:box_climb_solver}
\end{figure}

Fig.~\ref{fig:perc:box_climb_solver} provides insight into the solver during the motion performed with the knee collisions included. The four peaks in the cost function show the effect of the collision avoidance penalty when the legs are close to the obstacle during the step up and step down. Most of the time, the step obtained from the QP subproblem is accepted by the line-search with the full stepsize of $1.0$. However, between \SI{7}{} and \SI{8}{\second} the stepsize is decreased to prevent the constraint violation from further rising. This happens when the front legs step down the box and are close to collision. In those cases, the collision avoidance penalty is highly nonlinear, and the line-search is required to maintain the right balance between cost decrease and constraint satisfaction. We note that the line-search condition for low constraint violation is typically not achieved when using only one iteration per MPC problem.

\subsubsection{Model selection}
\begin{figure}[!tb]
\centering
\includegraphics[width=\columnwidth]{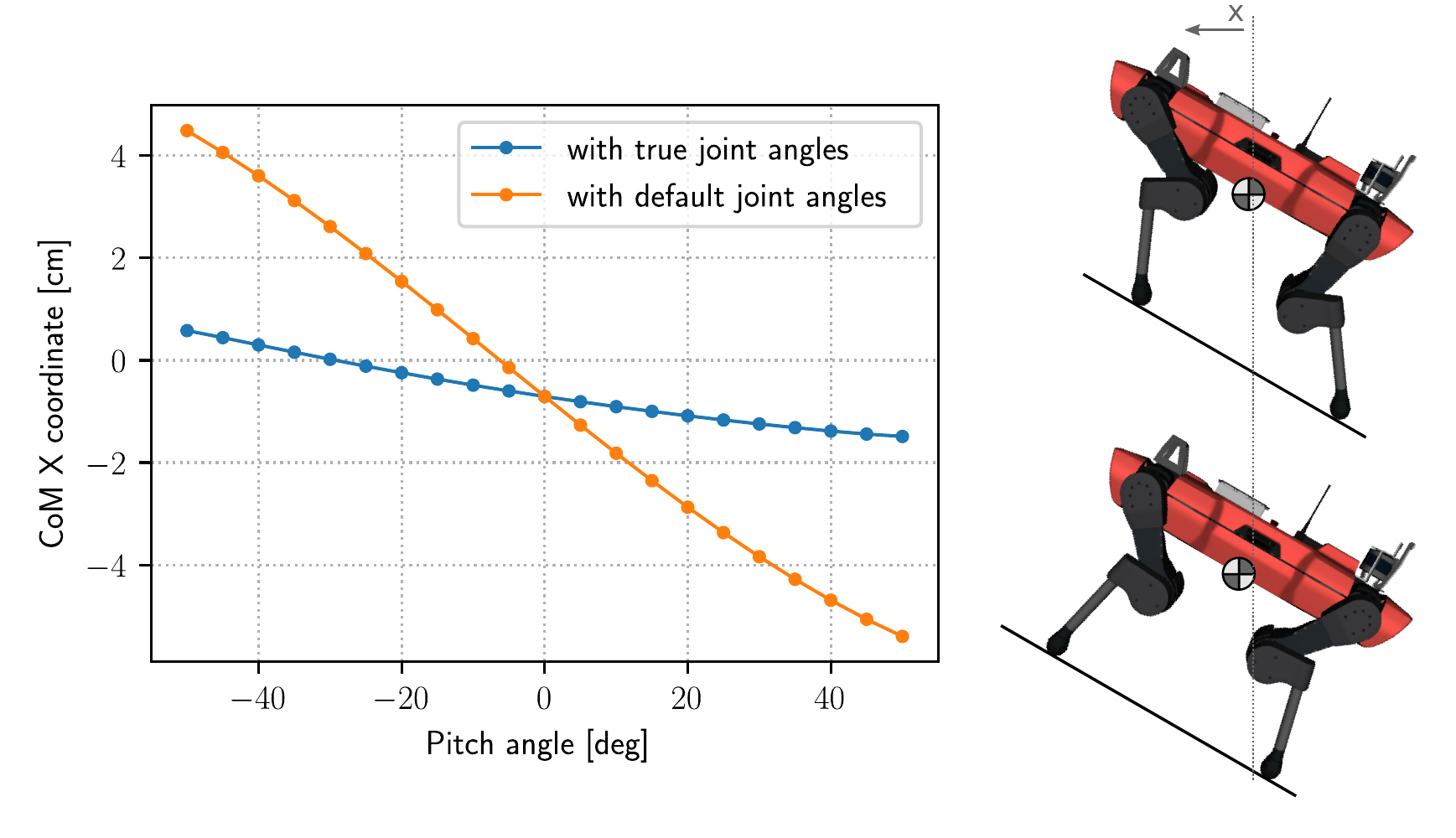}
\caption{\new{The location of the center of mass (CoM) in heading direction for various torso pitch angles. The first set of CoM locations is evaluated with the \textit{true} joint angles, which are obtained when aligning the legs with the gravity direction as in the top image. This corresponds to the reference in section \ref{sect:perc:reference_generation}, which is tracked by the MPC. The second set of CoM locations is evaluated for the \textit{default} joint angles, shown in the bottom image, as assumed by the SRBD model.}}
\label{fig:perc:com_comparison}
\end{figure}

In the same scenario, we compare the performance of the proposed dynamics for the base with those of the commonly used single rigid body dynamics (SRBD). To be precise, the torso dynamics in \eqref{eq:perc:torso_dynamics} are evaluated at a constant nominal joint configuration and with zero joint velocities, while the rest of the controller remains identical. When using the SRBD, the model does not describe the backward shift in the center of mass location caused by the leg configuration.  The result is that the controller with the SRBD model has a persisting bias that makes the robot almost tip over during the step up. \new{This model error is quantified in Fig.~\ref{fig:perc:com_comparison}. At \SI{30}{\degree} pitch angle, there is a center of mass error of \SI{2.6}{\centi\meter}, resulting in a bias of \SI{13.3}{\newton\meter} at the base frame. For reference, this is equivalent to an unmodelled payload of \SI{3.6}{\kilo\gram} at the tip of the robot.} The proposed model fully describes the change in inertia and center of mass location and therefore does not have any issue to predict the state trajectory during the step up motion. 

%\begin{figure}[!tb]
%    \centering
%    \begin{minipage}{.49\columnwidth}
%        \centering
%        \includegraphics[width=\linewidth,trim=400 200 500 150,clip]{figures/box_climb_srbd.png}
%    \end{minipage}
%    \begin{minipage}{.49\columnwidth}
%        \centering
%        \includegraphics[width=\linewidth,trim=400 200 500 150,clip]{figures/box_climb_full_model.png}
%    \end{minipage}
%        \caption{ANYmal after stepping up a box of \SI{30}{\centi\meter}. Left: using the SRBD model. Right: using the proposed system dynamics. The simplifications in the SRBD model cause a significant disturbance during the step up motion.}
%        \label{fig:perc:box_climb_srbd}
%\end{figure}

\begin{figure*}[!bt]
\centering
\includegraphics[width=\linewidth,trim={0 0 0 0},clip]{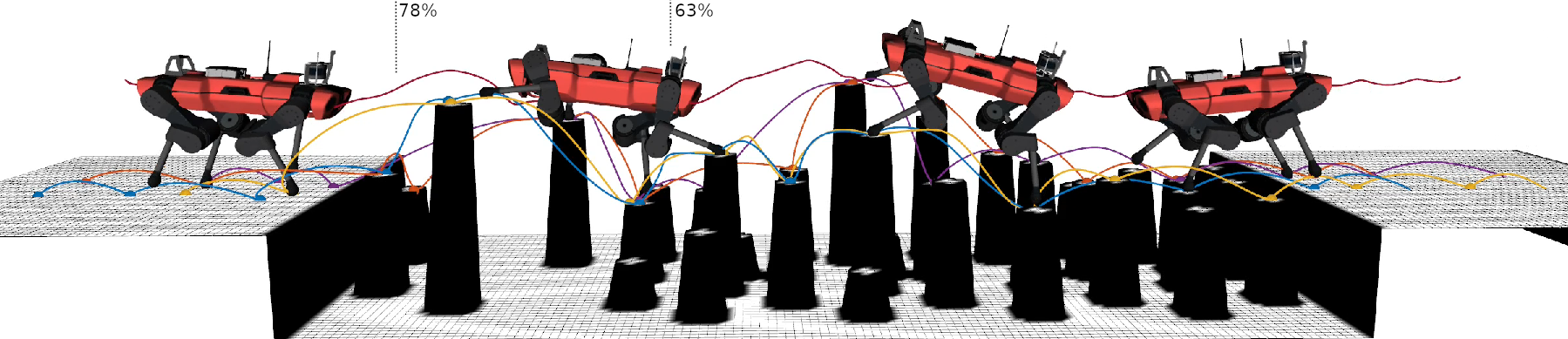}
\caption{\new{ANYmal traversing stepping stones in simulation (right to left). The resulting state trajectories for feet and torso, and the snapshots are shown for a traversal with the multiple-shooting solver and a trotting gait at \SI{0.75}{\meter\per\second}. The marked \SI{63}{\percent} and \SI{78}{\percent} locations indicate where the alternative solver, iLQR, diverges for \SI{0.5}{\meter\per\second} and \SI{0.75}{\meter\per\second}, respectively.}}
\label{fig:perc:stepping_stones_sim}
\end{figure*}

\new{
\subsubsection{Solver Comparison}
To motivate our choice to implement a multiple-shooting solver and move away from the DDP-based methods used in previous work, we compare both approaches on flat terrain and the stepping stone scenario shown in Fig.~\ref{fig:perc:stepping_stones_sim}. In particular, we compare against iLQR~\cite{tassa2012synthesis} and implement it with the same constraint projection, line-search, and the Riccati Backward pass of HPIPM as described in section~\ref{sect:perc:numerical_optimization}. The key difference between the algorithms lies in the update step. For multiple-shooting, we update both state and inputs directly: $\vu^+_k = \vu_k + \alpha\delta\vu_k$, $\vx^+_k = \vx_k + \alpha\delta\vx_k$. In contrast, iLQR proceeds with a line-search over closed-loop nonlinear \textit{rollouts} of the dynamics:
\begin{align}
\vu^+_k &=  \vu_k + \alpha \vk_k + \vK_k \left(\vx^+_k - \vx_k \right), \\
\vx^+_{k+1} &= \vf^d_k(\vx^+_k, \vu^+_k),  \quad\qquad \vx^+_0 = \hat{\vx},
\end{align}
where $\vK_k$ is the optimal feedback gain obtained from the Riccati Backward pass and $\vk_k = \delta \vu_k - \vK_k \delta \vx_k $ is the control update. Due to this inherently single-threaded process, each line-search for iLQR takes four times as long as for the multi-threaded multiple-shooting. However, note that with the hybrid multiple-shooting-iLQR variants in \cite{giftthaler2018family} this difference vanishes.

Table~\ref{tab:perc:solver_comparison} reports the solvers' average cost, dynamics constraint violation, and equality constraint violation for a trotting gait in several scenarios.   As a baseline, we run the multiple-shooting solver until convergence (with a maximum of 50 iterations) instead of real-time iteration. To test the MPC in isolation, we use the MPC dynamics as the simulator and apply the MPC input directly. Because of the nonlinear rollouts of iLQR, dynamics constraints are always satisfied, and iLQR, therefore, has the edge over multiple-shooting on this metric. However, as the scenario gets more complex and the optimization problem becomes harder, there is a point where the forward rollout of iLQR is unstable and diverges. For the scenario shown in Fig.~\ref{fig:perc:stepping_stones_sim}, this happens in the place where the robot is forced to take a big leap at the \SI{63}{\percent} mark and at the \SI{78}{\percent} mark where the hind leg is close to singularity as the robot steps down. The continuous time variant SLQ~\cite{farshidian2017efficient} fails in similar ways. These failure cases are sudden, unpredictable, and happen regularly when testing on hardware, where imperfect elevation maps, real dynamics, and disturbances add to the challenge. The absence of long horizon rollouts in the multiple-shooting approach makes it more robust and better suited for the scenarios shown in this work. For cases where both solvers are stable, we find that the small dynamics violation left with multiple-shooting in a real-time iteration setting does not translate to any practical performance difference on hardware.
% If such cases do appear in the future, hybrid DDP-multiple-shooting variants could be considered~\cite{mastalli2020crocoddyl,giftthaler2018family}.
Finally, even for the most challenging scenario, multiple-shooting with real-time iteration remains within \SI{10}{\percent} cost of the baseline.
}

\begin{table}[tb]
\caption{Solver comparison on flat terrain and stepping stones. The Baseline iterates until convergence instead of using real-time iteration.}
\label{tab:perc:solver_comparison}
\begin{tabular}{lrrr}
\hline
\multicolumn{1}{l}{} & \multicolumn{1}{c}{Baseline}\rule{0pt}{14pt} & \multicolumn{1}{c}{\begin{tabular}[c]{@{}c@{}}Multiple\\ shooting\end{tabular}}& \multicolumn{1}{c}{iLQR}    \\ 
\hline 
\multicolumn{1}{l}{\textit{Flat - \SI{0.50}{\meter/\second}}}\rule{0pt}{12pt}  &      &        &     \\
Cost    		& $ 54.19$   			& $ 54.17$     			& $54.18 $       		       		\\
Dynamics Constr.    & $ 1.70\times 10^{-7}$  & $3.41\times 10^{-3}$   & $0.0 $    \\
Equality Constr.      & $ 2.28\times 10^{-6}$  & $3.51\times 10^{-3}$	& $3.51 \times 10^{-3}$     \\
\multicolumn{1}{l}{\textit{Stones - \SI{0.25}{\meter/\second}}}\rule{0pt}{12pt}  &      &        &     \\
Cost    		& $ 151.92$   			& $ 156.22$     			& $ 156.69$       		       		\\
Dynamics Constr.    & $ 3.58 \times 10^{-5}$  & $1.01\times 10^{-2}$            & $0.0$    \\
Equality Constr.      & $ 7.24 \times 10^{-4}$  & $2.28\times 10^{-2}$	& $2.14\times 10^{-2}$     \\ 
\multicolumn{1}{l}{\textit{Stones - \SI{0.50}{\meter/\second}}}\rule{0pt}{12pt}   &      &        &     \\
Cost    		& $ 155.06 $   			& $ 165.72 $     			& \multirow{3}{*}{\begin{tabular}[c]{@{}c@{}}diverged at $78\%$ \\ scenario progress\end{tabular}}	\\
Dynamics Constr.    & $ 2.18\times 10^{-5}$  & $1.53\times 10^{-2}$            &     \\
Equality Constr.      & $ 3.93\times 10^{-4}$  & $3.82\times 10^{-2}$	&      \\ 
\multicolumn{1}{l}{\textit{Stones - \SI{0.75}{\meter/\second}}}\rule{0pt}{12pt}   &      &        &     \\
Cost    		& $ 199.49$   			& $215.98 $     			& \multirow{3}{*}{\begin{tabular}[c]{@{}c@{}}diverged at $63\%$ \\ scenario progress\end{tabular}}        		       		\\
Dynamics Constr.    & $ 1.20\times 10^{-4}$  & $2.36\times 10^{-2}$            &     \\
Equality Constr.      & $ 1.29\times 10^{-3}$  & $5.61\times 10^{-2}$	&      \\ 
\hline
\end{tabular}
\end{table}

\subsubsection{Contact feedback} 
\begin{figure}[!tb]
\centering
\includegraphics[width=0.7\columnwidth]{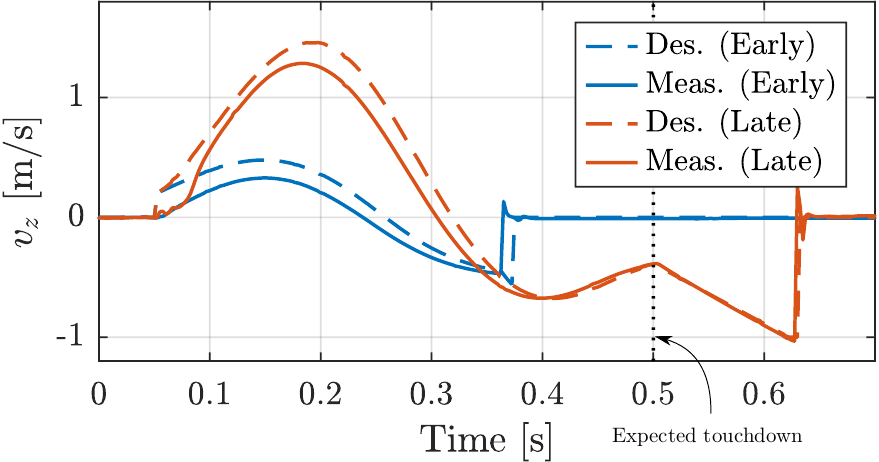}
\caption{Desired and measured vertical foot velocity for the early and late touchdown scenarios shown in the accompanying video. The vertical line at \SI{0.5}{\second} indicated the planned touchdown time.}
\label{fig:perc:touchdown_vel}
\end{figure}

The reactive behavior under a mismatch in planned and sensed contact information is shown in the accompanying video. First, the sensed terrain is set to be \SI{10}{\centi\meter} above the actual terrain, causing a late touchdown. Afterward, the sensed terrain is set \SI{5}{\centi\meter} below the actual terrain, causing an early touchdown. The resulting vertical foot velocity for both cases is overlayed and plotted in Fig.~\ref{fig:perc:touchdown_vel}. For the case of a late touchdown, the reactive downward accelerating trajectory is triggered as soon as it is sensed that contact is absent. For the early touchdown case, there is a short delay in detecting that contact has happened, but once contact is detected, the measured contact is included in the MPC and the new trajectory is immediately replanned from the sensed contact location.

\begin{figure*}[!bt]
\centering
\includegraphics[width=\linewidth,trim={250 310 200 290},clip]{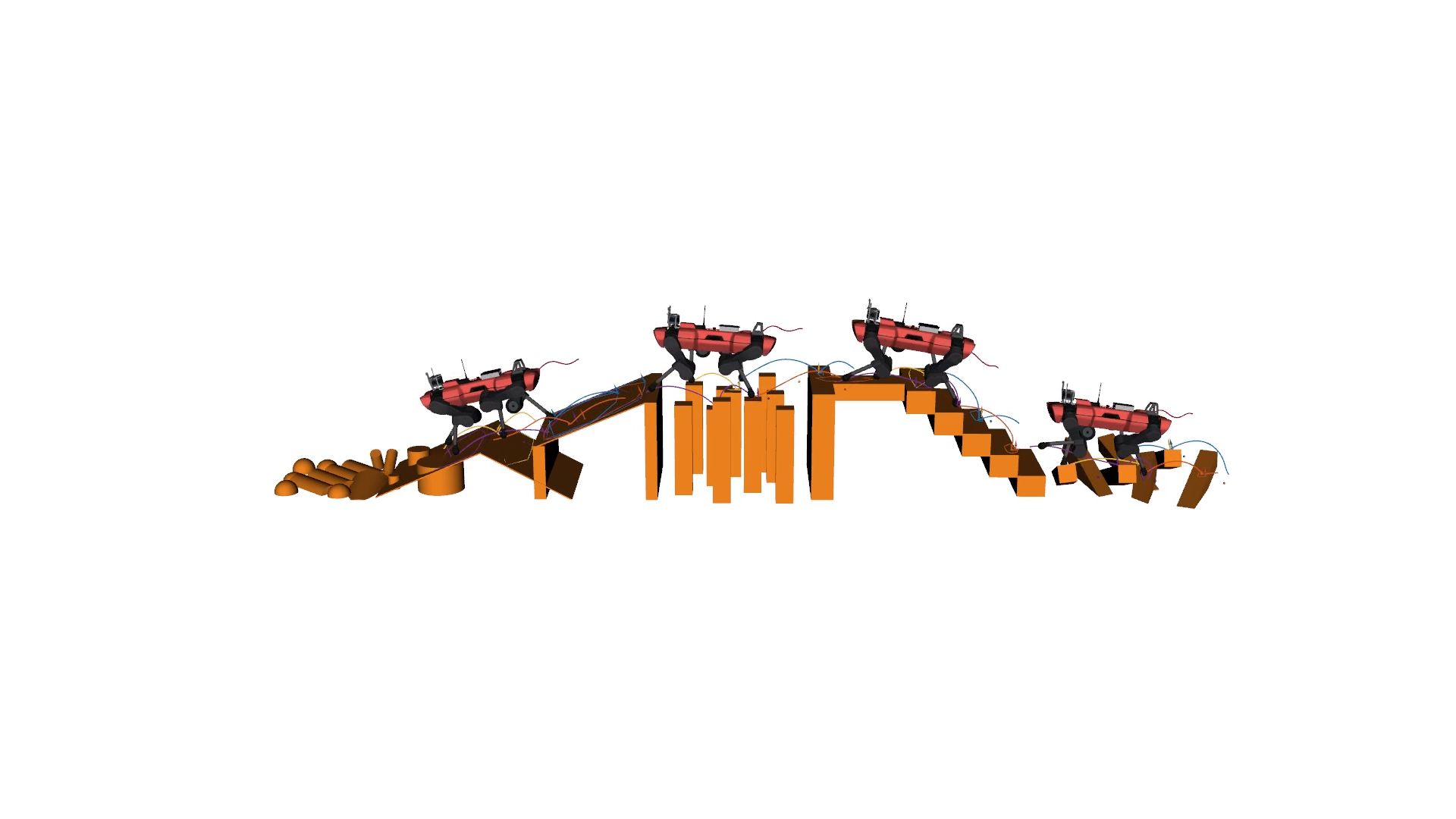}
\caption{ANYmal traversing an obstacle course in simulation \new{(left to right)}. Snapshots are shown for a traversal with a trotting gait at \SI{0.8}{\meter\per\second}. The MPC predictions are shown for each foot and for the torso center. For all contact phases within the horizon, the convex foot placements contraints are visualized.}
\label{fig:perc:obstacle_course}
\end{figure*}

\subsubsection{Stairs}
The generality of the approach with respect to the gait pattern is demonstrated in the accompanying video by executing a trot at \SI{0.25}{\meter\per\second}, a pace at \SI{0.3}{\meter\per\second}, a dynamic walk at \SI{0.25}{\meter\per\second}, and a static walk at \SI{0.2}{\meter\per\second} on a stairs with \SI{18.5}{\centi\meter} rise and  \SI{24}{\centi\meter} run. Depending on the particular gait pattern and commanded velocity the method autonomously decides to progress, repeat, or skip a step. Note that there are no parameters or control modes specific to the gait or the stair climbing scenario. All motions emerge automatically from the optimization of the formulated costs and constraints.

\subsubsection{Obstacle course} The controller is given a constant forward velocity command on a series of slopes, gaps, stepping stones, and other rough terrains. We traverse the terrain with a pace at \SI{0.4}{\meter\per\second}, and a fast trotting gait with flight phase at \SI{0.8}{\meter\per\second}. Fig.~\ref{fig:perc:obstacle_course} shows the obstacle course and snapshots of the traversal with the fast trot. The supplemental video shows the planned trajectories for the feet together with the convex foothold constraints. In the right side of the screen, a front view is shown together with the elevation map and plane segmentation below. The slower gaits used in the previous section are able to complete the scenario as well, but their video is excluded as they take long to reach the end. 

Finally, a transverse gallop gait is demonstrated on a series of gaps. Due to the torque limitations of the system and friction limits up the slope, this gait is not feasible on the more complex obstacle course.

\subsubsection{Comparison against RL} 
We compare our method against a perceptive RL-based controller~\cite{miki2022learning} in the same obstacle course. We adapt the gait pattern of our controller to match the nominal gait used by the learned controller. The video shows that the learning-based controller can cross the unstructured terrain at the beginning and end of the obstacle course. However, it fails to use the perceptive information fully and falls between the stepping stones when starting from the left and off the narrow passage when starting from the right. \new{While the RL controller was not specifically trained on stepping stones,} this experiment highlights that current RL-based locomotion results in primarily reactive policies and struggles with precise coordination and planning over longer horizons. In contrast, using a model and online optimization along a horizon makes our proposed method generalize naturally to these more challenging terrains.

\subsection{Hardware}
\label{sect:perc:results_hardware}

\begin{figure*}[!tb]
\centering
\begin{minipage}{\linewidth}
\centering
\includegraphics[width=\linewidth,trim={0 60 0 180},clip]{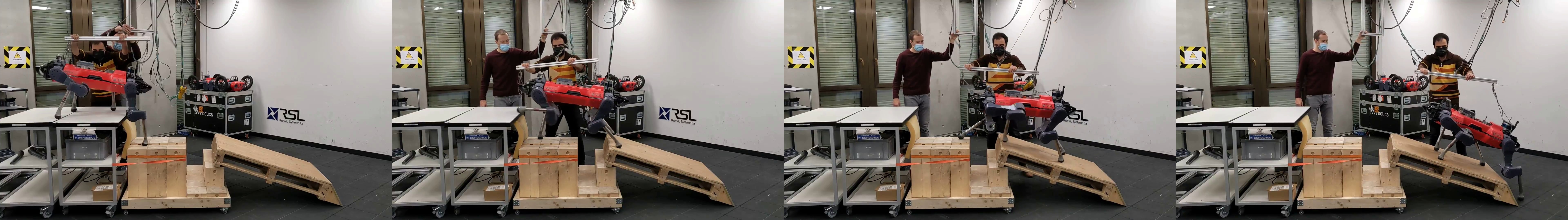}
\end{minipage}
\begin{minipage}{\linewidth}
\centering
\includegraphics[width=\linewidth,trim={0 60 0 180},clip]{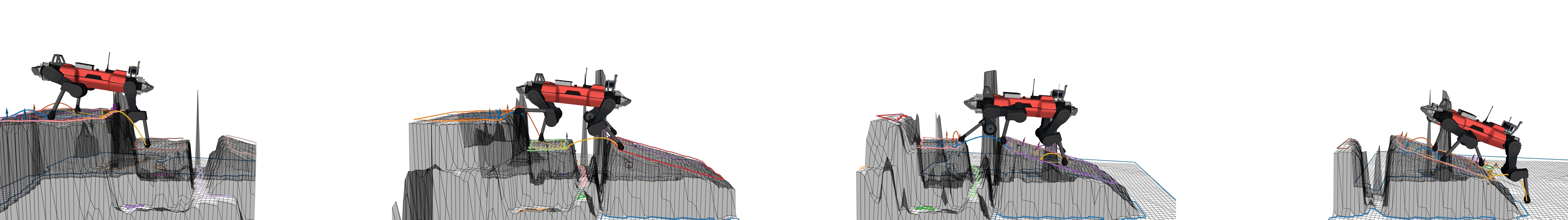}
\end{minipage}
\caption{Hardware experiment where ANYmal traverses a ramp, gap, and large step (from right to left). The bottom row shows the filtered elevation map, the foot trajectories over the MPC horizon, and the convex foothold constraints.}
\label{fig:perc:tableclimbing}
\end{figure*}

\begin{figure}[!tb]
\centering
\begin{minipage}{0.49\linewidth}
\centering
\includegraphics[width=\linewidth,trim={100 60 200 0},clip]{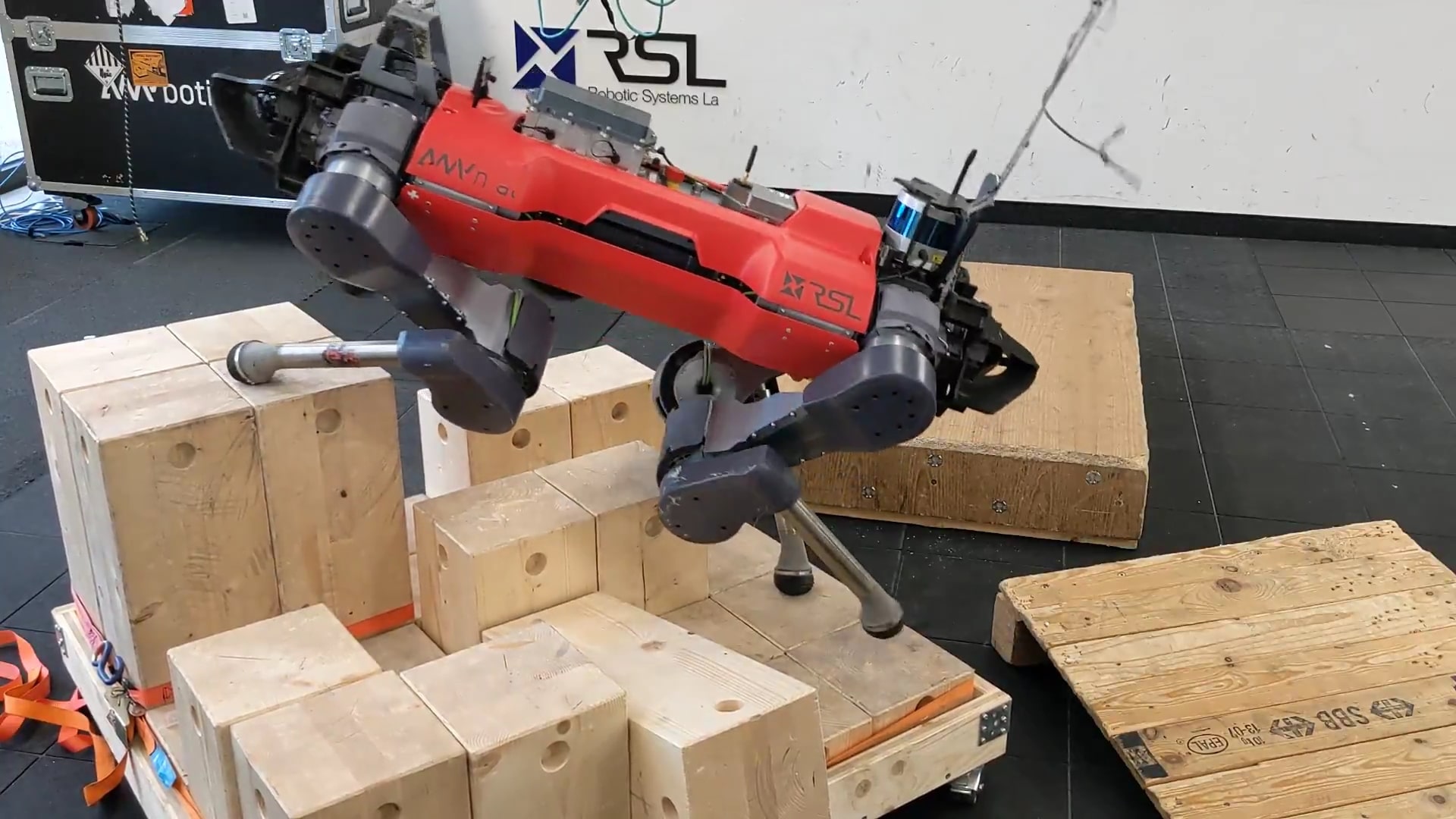}
\end{minipage}%
\begin{minipage}{0.49\linewidth}
\centering
\includegraphics[width=\linewidth,trim={150 150 300 0},clip]{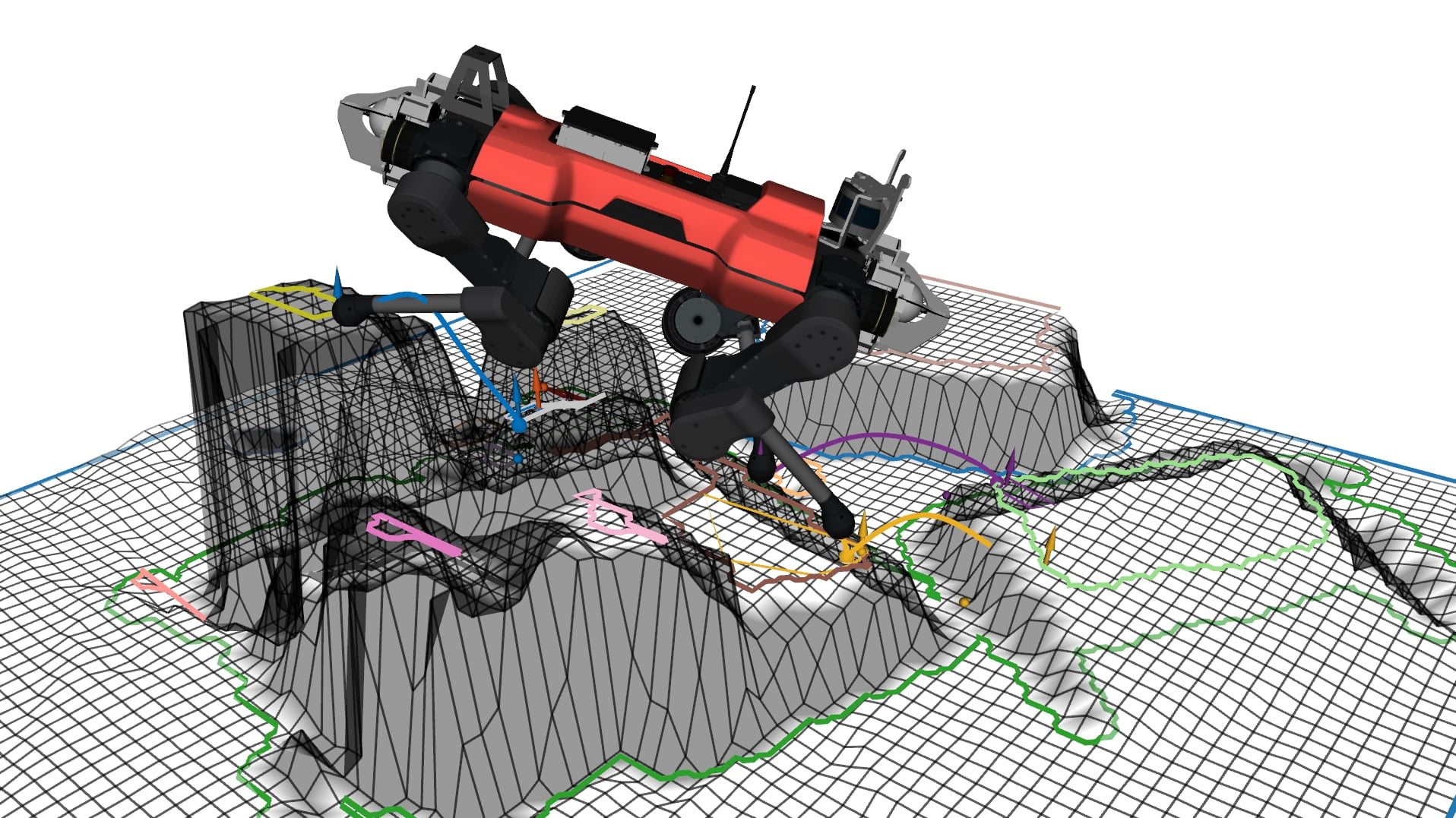}
\end{minipage}
\caption{Hardware experiment where ANYmal walks on top of uneven stepping stones. Each wooden block has an area of \SI{20}{}x\SI{20}{\centi\meter} and each level of stepping stones is \SI{20}{\centi\meter} higher than the previous one. The right image shows the filtered elevation map, the foot trajectories over the MPC horizon, and the convex foothold constraints.}
\label{fig:perc:steppingstones}
\end{figure}

\subsubsection{Obstacle course}
The obstacle course simulation experiment is recreated on hardware in two separate experiments. First, we tested a sequence of a ramp, gap, and high step as shown in Fig.~\ref{fig:perc:tableclimbing}. During the middle section of this experiment, the robot faces all challenges simultaneously: While the front legs are stepping up to the final platform, the hind legs are still dealing with the ramp and gap. In a second scenario, the robot is walking on a set of uneven stepping stones, as shown in Fig.~\ref{fig:perc:steppingstones}. The main challenge here is that the planes on the stepping stones are small and do not leave much room for the MPC to optimize the footholds. We found that in this scenario, the inclusion of the kinematics and reactive foothold offset during the plane selection as described in section~\ref{sect:perc:reference_generation} are important. A remaining challenge here is that our plane segmentation does not consider consistency over time. In some cases, the small foothold regions on top of stepping stones might appear and disappear as feasible candidates. The supplemental video shows how in this case the planned foot trajectory can fail, and the reactive contact regaining is required to save the robot.

\begin{table}[!tb]
\caption{Computation times per map update and MPC iteration}
\label{tab:perc:computation_times}
\centering
\begin{tabular}{r|rrr}
            & Mean [\SI{}{\milli\second}] & Max [\SI{}{\milli\second}] \\ \hline
Classification \& Segmentation	& 38.8 &  76.6 \\
Signed distance field	& 1.3 &  7.6 \\ \hline
LQ approximation  & 3.6  & 6.2 \\
QP solve    & 2.7  & 4.4 \\
Line-search &  0.3 & 0.9 \\ \hline
MPC iteration &  6.6 & 9.8 \\ 

\end{tabular}
\end{table}

Computation times are reported in Table~\ref{tab:perc:computation_times}. Per map update, most time is spent on terrain classification and plane segmentation. More specifically, the RANSAC refinement takes the most time and can cause a high worst-case computation due to its sampling-based nature. On average, the perception pipeline is able to keep up with the \SI{20}{\hertz} map updates.

For the MPC computation time, the `LQ approximation` contains the parallel computation of the linear-quadratic model and equality constraint projection (Algorithm~\ref{alg:SQP}, line \ref{alg:line:dt} till \ref{alg:line:proj}). `QP solve` contains the solution of the QP and the back substitution of the solution (Algorithm~\ref{alg:SQP}, line \ref{alg:line:hpipmsolve} and \ref{alg:line:backproj}). Despite the parallelization across four cores, evaluating the model takes the majority of the time, with the single core solving of the QP in second place. On average, the total computation time is sufficient for the desired update rate of \SI{100}{\hertz}. The worst-case computation times are rare, and we hypothesize that they are mainly caused by variance in the scheduling of the numerous parallel processes on the robot. For the line-search, the relatively high maximum computation time is attained when several steps are rejected, and the costs and constraints need to be recomputed.

\begin{figure}[!tb]
\centering
\begin{minipage}{0.49\linewidth}
\centering
\includegraphics[width=\linewidth,trim={400 100 300 0},clip]{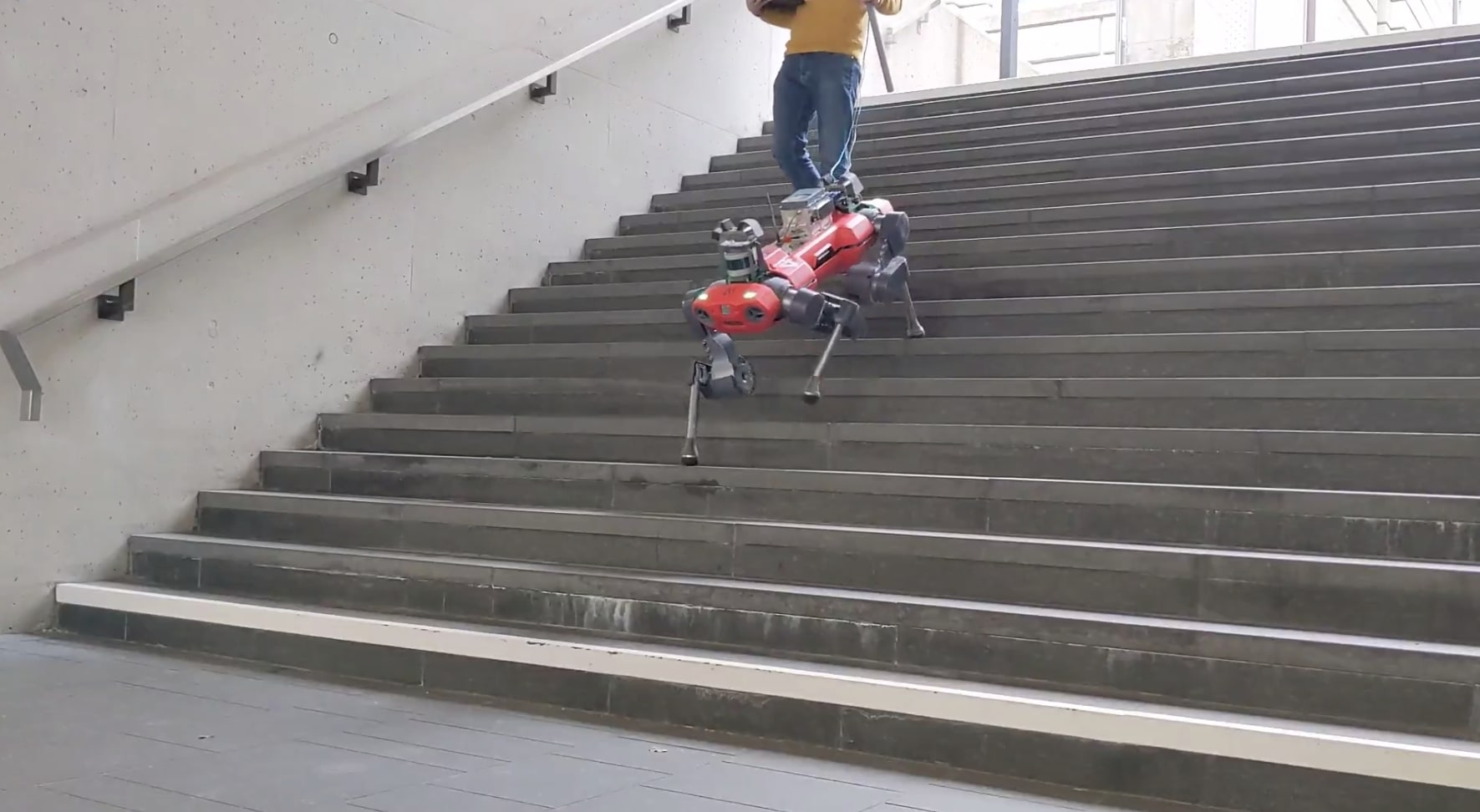}
\end{minipage}%
\begin{minipage}{0.49\linewidth}
\centering
\includegraphics[width=\linewidth,trim={300 55 400 0},clip]{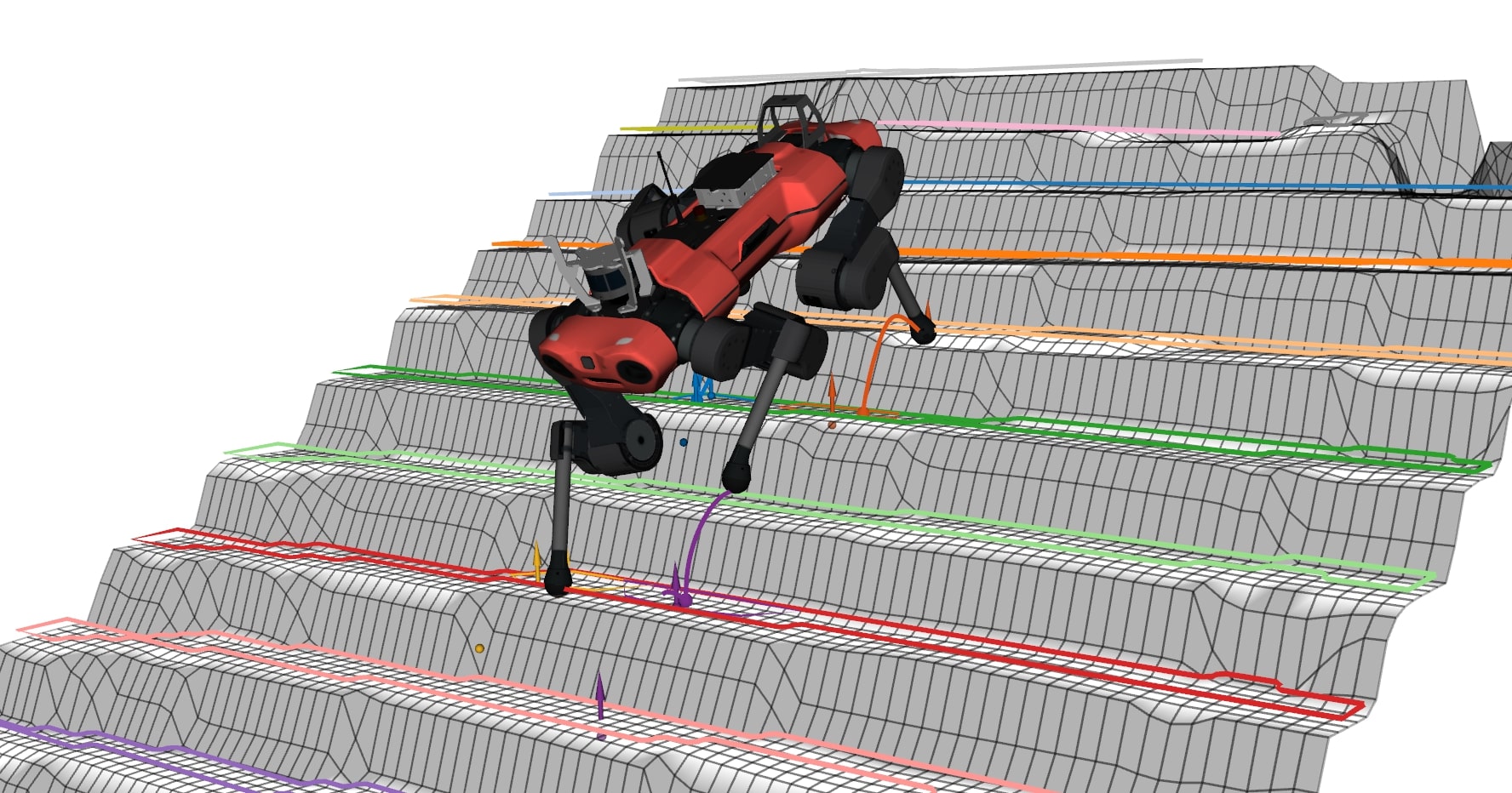}
\end{minipage}
\caption{Hardware experiment where ANYmal walks up and down outdoor stairs with a \SI{16}{\centi\meter} rise and \SI{29.5}{\centi\meter} run. The right image shows the filtered elevation map, the foot trajectories over the MPC horizon, and the convex foothold constraints.}
\label{fig:perc:outdoor_stairs}
\end{figure}

\subsubsection{Stairs} 
\begin{figure}[!tb]
\centering
\includegraphics[width=1.0\columnwidth]{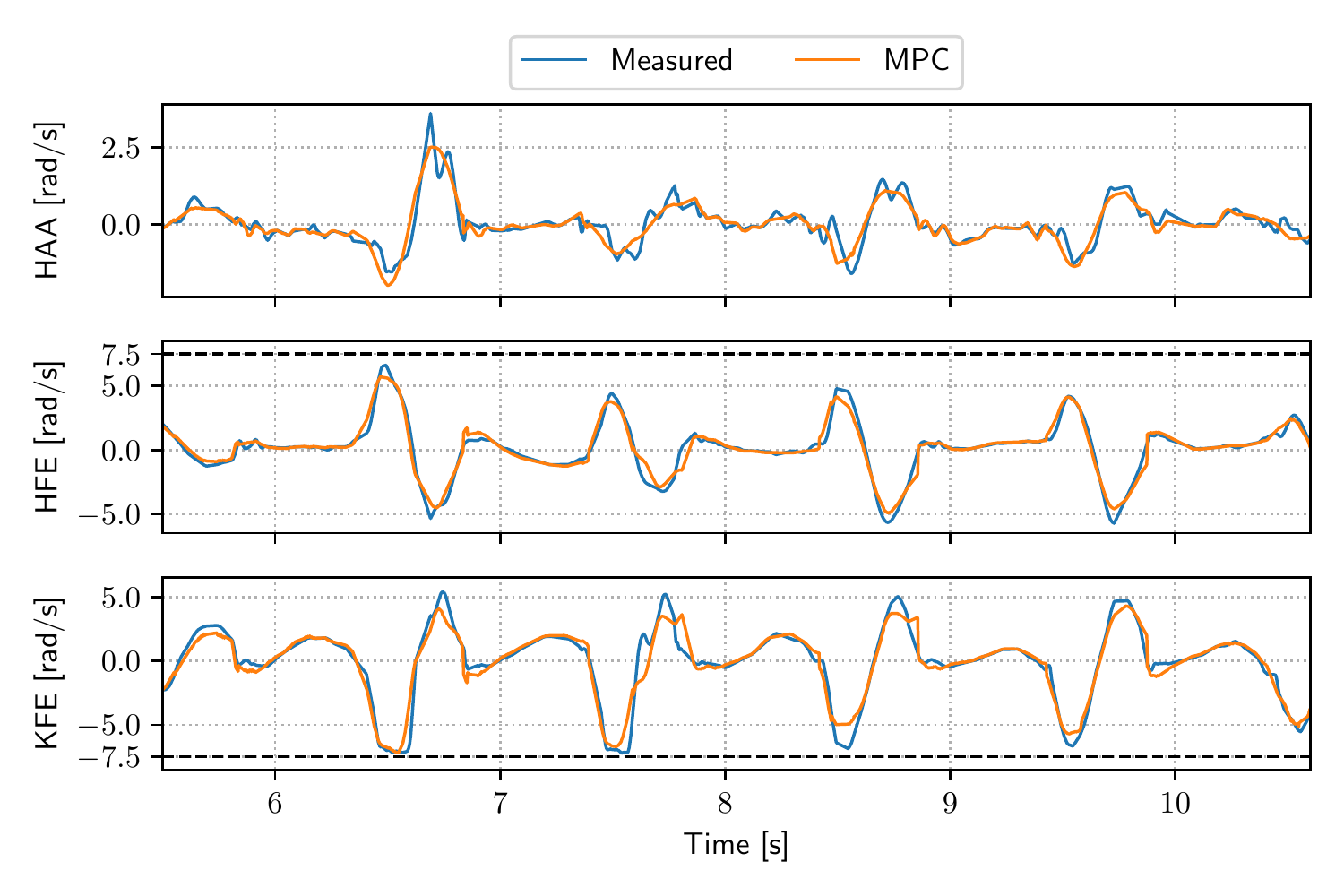}
\caption{\new{Measured and MPC commanded joint velocities for the left front leg while walking up the stairs shown in Fig~\ref{fig:perc:outdoor_stairs}. All joints, Hip Abduction Aduction (HAA), Hip Flexion Extension (HFE), and Knee Flextion Extension (KFE), have a velocity limit of $\pm$\SI{7.5}{\radian\per\second}.}}
\label{fig:perc:stairs_velocities}
\end{figure}

\begin{figure}[!tb]
\centering
\includegraphics[width=1.0\columnwidth]{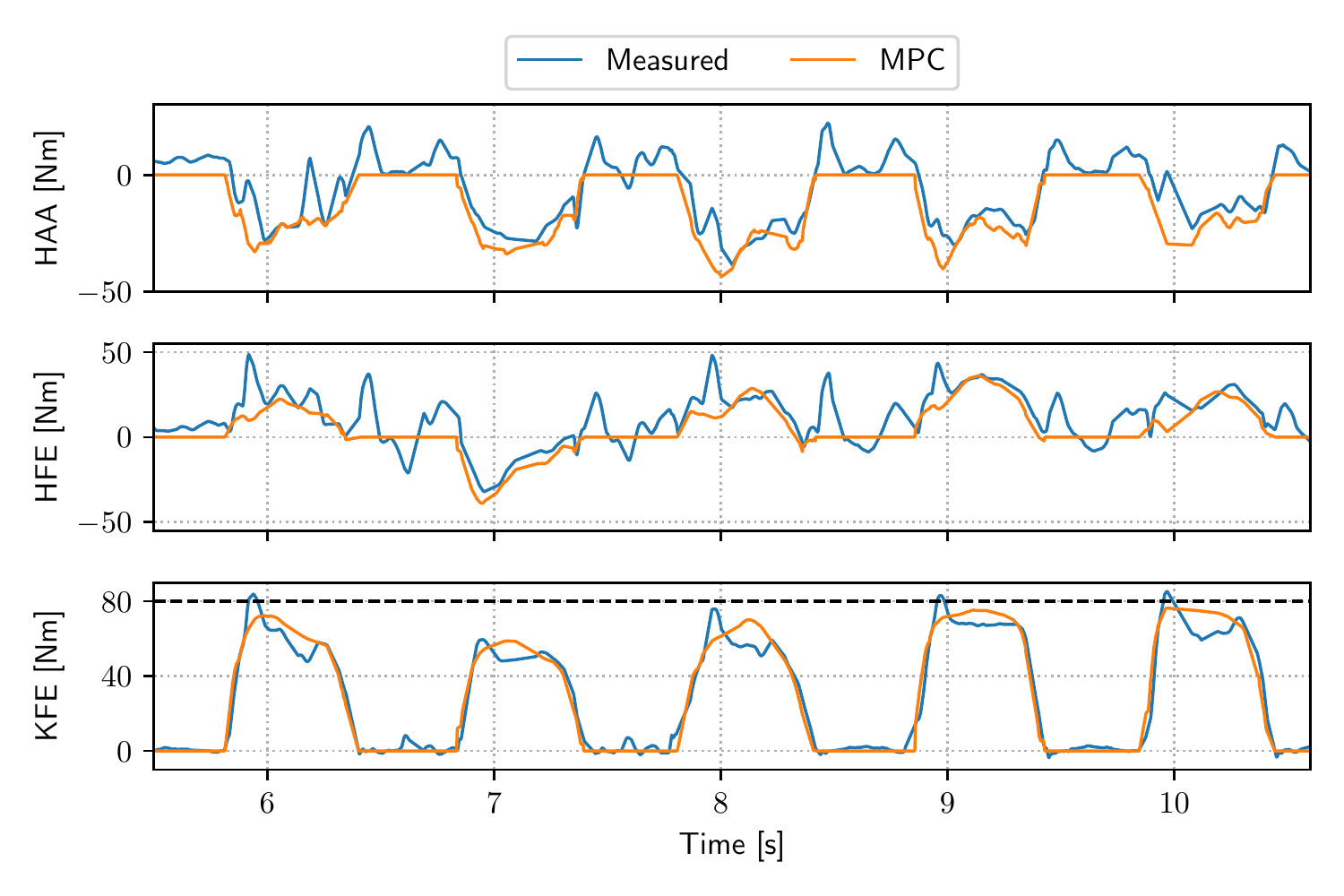}
\caption{\new{Measured torque and approximated torque within the MPC formulation $(\vtau_i = \vJ^\top_{j,i} \vlambda_i)$ for the left front leg while walking up the stairs shown in Fig~\ref{fig:perc:outdoor_stairs}. All joints, Hip Abduction Aduction (HAA), Hip Flexion Extension (HFE), and Knee Flextion Extension (KFE), have a torque limit of $\pm$\SI{80}{\newton\meter}.}}
\label{fig:perc:stairs_torques}
\end{figure}

We validate the stair climbing capabilities on 2-step indoor stairs and on outdoor stairs. Fig.~\ref{fig:perc:outdoor_stairs} shows the robot on its way down the outdoor stairs. For these experiments, we obtain the elevation map from~\cite{hoeller2022neural}. With its learning-based approach, it provides a high quality estimate of the structure underneath the robot. \new{Note that this module only replaces the source of the elevation map in Fig.~\ref{fig:perc:perception_overview}, and does not change the rest of our perception pipeline. Fig.~\ref{fig:perc:stairs_velocities} and \ref{fig:perc:stairs_torques} show the measured joint velocities and torques alongside the same quantities within the MPC solution for five strides of the robot walking up the stairs. The optimized MPC values are within the specified limits and close to the measured values.}

\subsection{Limitations}
\label{sect:perc:results_limitations}
A fundamental limitation in the proposed controller is that the gait pattern is externally given and only adapted during early and late touchdown. Strong adverse disturbances, for example, in the direction of a foot that will soon lift, can make the controller fail. A change in the stepping pattern could be a much better response in such cases. Together with the reactive behaviors during contact mismatch, which are currently hardcoded, we see the potential for reinforcement learning-based methods as a tracking controller to add to the robustness during execution.

Closely related to that, the current selection of the segmented plane and, therefore, the resulting foothold constraints happens independently for each leg. In some cases, this can lead to problems that could have been avoided if all legs were considered simultaneously. For example, while walking up the stairs sideways, all feet can end up on the same tread, leading to fragile support and potential self-collisions. \new{Similarly, the presented method targets local motion planning and control, and we should not expect global navigation behavior. The current approach will attempt to climb over gaps and obstacles if so commanded by the user and will not autonomously navigate around them.}

\begin{figure}[!tb]
\centering
\includegraphics[width=0.70\columnwidth,trim={0 0 0 0},clip]{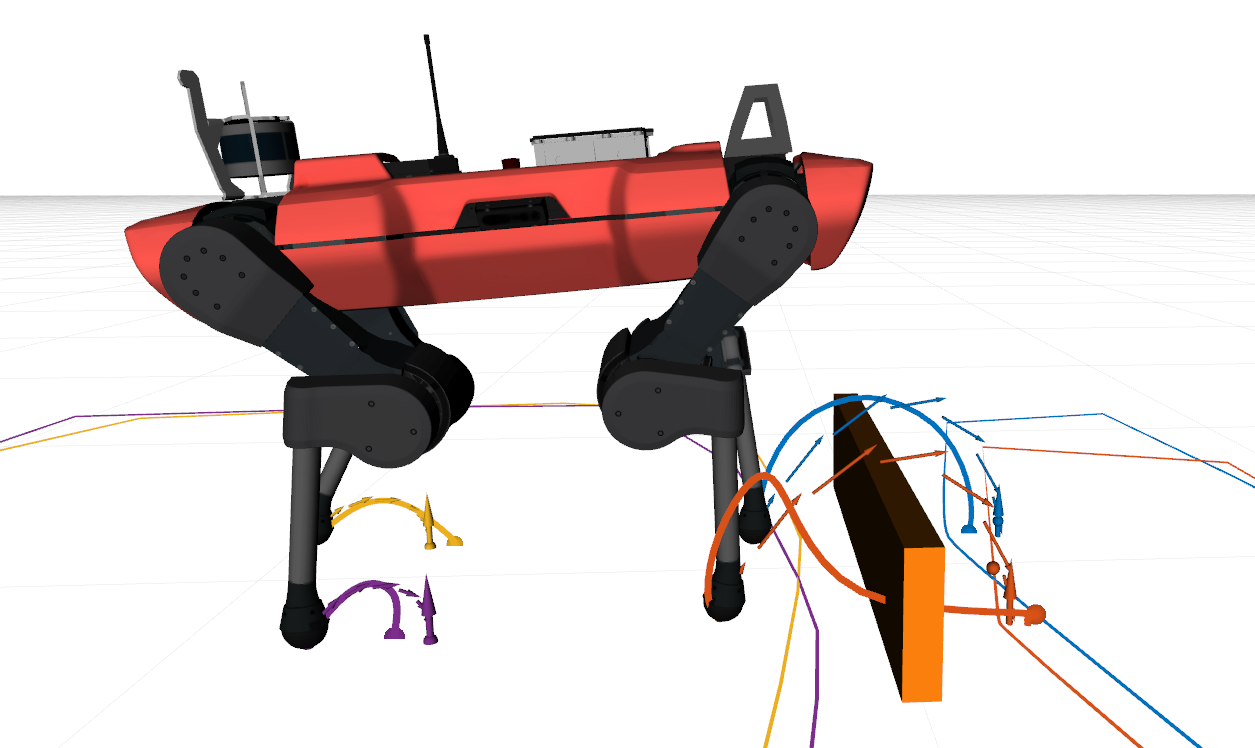}
\caption{Example of the MPC optimization being stuck inside a tall and thin structure of \SI{5}{\centi\meter} width and \SI{20}{\centi\meter} height. The feet reference trajectories used as part of the cost function are visualized as a sequence of arrows. }
\label{fig:perc:hurdles_stuck}
\end{figure}

As with all gradient-based methods for nonlinear optimization, local optima and infeasibility can be an issue. With the simplification of the terrain to convex foothold constraints and by using a heuristic reference motion in the cost function, we have aimed to minimize such problems. Still, we find that in the case of very thin and tall obstacles, the optimization can get stuck. Fig.~\ref{fig:perc:hurdles_stuck} shows an example where the foothold constraints lie behind the obstacle and the reference trajectory correctly clears the obstacle. Unfortunately, one of the feet in the MPC trajectory goes right through the obstacle. Because all SDF gradients are horizontal at that part of the obstacle, there is no strong local hint that the obstacle can be avoided. For future work, we can imagine detecting such a case and triggering a sampling-based recovery strategy to provide a new, collision-free initial guess. Alternatively, recent learning-based initialization could be employed \cite{melon2021receding,lembono2020learning}.

\new{Finally, we show a gallop and trot with flight phases at the end of the video. For these motions, the perceptive information is turned off, and the robot estimates the ground plane through a history of contact points. It demonstrates that the presented MPC is ready to express and stabilize these highly dynamic motions. Unfortunately, the elevation map is not usable due to artefacts from impacts and state estimation drift.}

\section{Conclusion}
\label{sect:perc:conclusion}
In this work, we proposed a controller capable of perceptive and dynamic locomotion in challenging terrain. By formulating perceptive foot placement constraints through a convex inner approximation of steppable terrain, we obtain a nonlinear MPC problem that can be solved reliably and efficiently with the presented numerical strategy. Steppability classification, plane segmentation, and an SDF are all precomputed and updated at \SI{20}{\hertz}. Asynchronously precomputing this information minimizes the time required for each MPC iteration and makes the approach real-time capable. Furthermore, by including the complete joint configuration in the system model, the method can simultaneously optimize foot placement, knee collision avoidance, and underactuated system dynamics. With this rich set of information encoded in the optimization, the approach discovers complex motions autonomously and generalizes across various gaits and terrains that require precise foot placement and whole-body coordination.

%% file: chapters/appendix.tex
% or


\section{Signed Distance Field Computation}
\label{sect:perc:sdf_appendix}

This section details how a signed distance field can be computed for a 2.5D elevation map. Consider the following general definition for the squared Euclidean distance between a point in space and the closest obstacle:
\begin{equation}
\begin{split}
    \mathcal{D}(x, y, z) = \min_{x', y', z'} \left[ \left( x - x' \right)^2 + \left( y - y' \right)^2 + \left( z - z' \right)^2 \right. \\ \left. + \, I(x', y', z')^{\phantom{2}}  \right],
\end{split}
\label{eq:perc:distanceDefinition}
\end{equation}
where $I(x', y', z')$ is an indicator function returning $0$ for an obstacle and $\infty$ for empty cells. 

As described in \cite{felzenszwalb2012distance}, a full 3D distance transform can be computed by consecutive distance transforms in each dimension of the grid, in arbitrary order. For the elevation map, the distance along the z-direction is trivial. Therefore, starting the algorithm with the z-direction simplifies the computation. First, \eqref{eq:perc:distanceDefinition} can be rewritten as follow,
\begin{align}
    \mathcal{D}(x, y, z) &= \min_{x', y'} \left[ \left( x - x' \right)^2 +  \left( y - y' \right)^2 + \right. \\ &\qquad\qquad \left. \min_{z'} \left[ \left( z - z' \right)^2 + I(x', y', z') \right]\right], \notag \\ 
    &= \min_{x', y'} \left[  \left( x - x' \right)^2 + \left( y - y' \right)^2 + f_z(x', y', z) \right],
    \label{eq:perc:2dSignedDistance}
\end{align}
where $f_z(x', y', z)$ is a function that returns for each horizontal position, the one-dimensional distance transform in z-direction. For an elevation map, this function has the following closed form solution at a given height $z$.
\begin{equation}
  f_z(x', y', z) = \begin{cases}
(z - h(x',y'))^2  & \text{if } z \geq h(x',y'), \\
0            & \text{otherwise},
\end{cases}
\end{equation}
where $h(x',y')$ denotes the evaluation of the elevation map.

\begin{figure}[!t]
\centering
\includegraphics[width=0.9\linewidth]{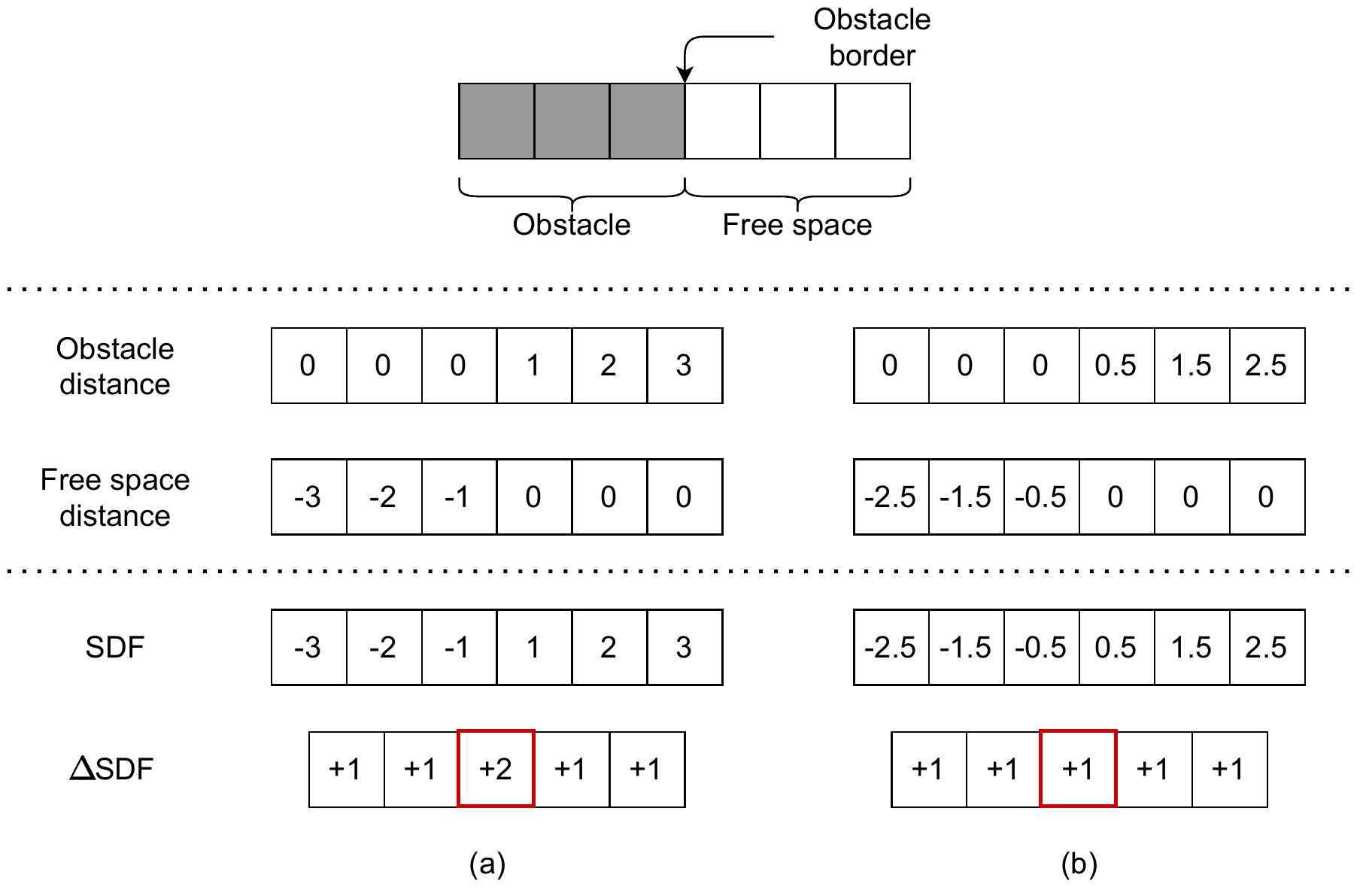}
\caption{1D example illustrating the effect of distance metric on the SDF. When taking the Euclidean distance between cell centers in (a), the SDF will have a discontinuous gradient across the obstacle border. Taking the distance between cell center and the border of an occupied / free cell as in (b), avoids this issue.}
\label{fig:perc:pixel_distance}
\end{figure}

The same idea can be used to compute the distance to obstacle free space and obtain the negative valued part of the SDF. Adding both distances together provides the full SDF and gradients are computed by finite differences between layers, columns, and rows. However, naively taking the Euclidean distance between cell centers as the minimization of \eqref{eq:perc:2dSignedDistance} leads to incorrect values around obstacle borders, as illustrated in Fig.~\ref{fig:perc:pixel_distance}. We need to account for the fact that the obstacle border is located between cells, not at the cell locations themselves. This can be resolved by adapting \eqref{eq:perc:2dSignedDistance} to account for the discrete nature of the problem.
\begin{equation}
    \mathcal{D}(x, y, z) = \min_{ \{x', y' \} \in \mathcal{M}} \left[  d \left( x, x'\right) + d \left( y, y' \right) + f_z(x', y', z) \right],
    \label{eq:perc:2dSignedDistanceDiscrete}
\end{equation}
where $\{x', y'\} \in \mathcal{M}$ now explicitly shows that we only minimize over the discrete cells contained in the map, and $d(\cdot, \cdot)$ is a function that returns the squared distance between the center of one cell and the border of another:
\begin{equation}
  d(x, x') = \begin{cases}
\left(|x - x'| - 0.5 r\right)^2  & \text{if } x \neq x', \\
0            & \text{otherwise},
\end{cases}
\end{equation}
where $r$ is the resolution of the map. The distance transforms can now be computed based on \eqref{eq:perc:2dSignedDistanceDiscrete}, for each height in parallel, with the 2D version of the algorithm described in \cite{felzenszwalb2012distance}.

%% file: chapters/acknowledgement.tex
% \section*{Acknowledgment}


%% file: chapters/biography.tex

\begin{IEEEbiography}[{\includegraphics[width=1in,height=1.25in,clip,keepaspectratio]{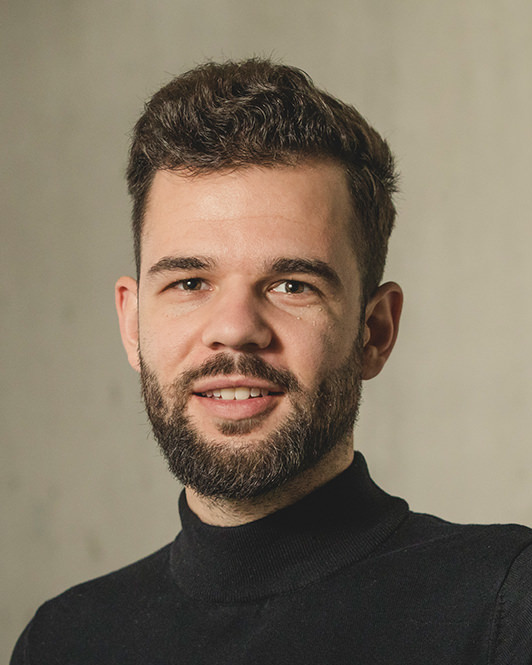}}]{Ruben Grandia}
received his B.Sc. in Aerospace Engineering from TU Delft, the Netherlands, in 2014, and his M.Sc. degree in Robotics, Systems, and Control from ETH Zurich, Switzerland, in 2017.
He is currently working toward the Ph.D. degree at the Robotic Systems Lab at ETH Zurich, under the supervision of Prof. M. Hutter. 
His research interests include nonlinear optimal control and its application to dynamic mobile robots.
\end{IEEEbiography}

\vfill

\begin{IEEEbiography}[{\includegraphics[width=1in,height=1.25in,clip,keepaspectratio]{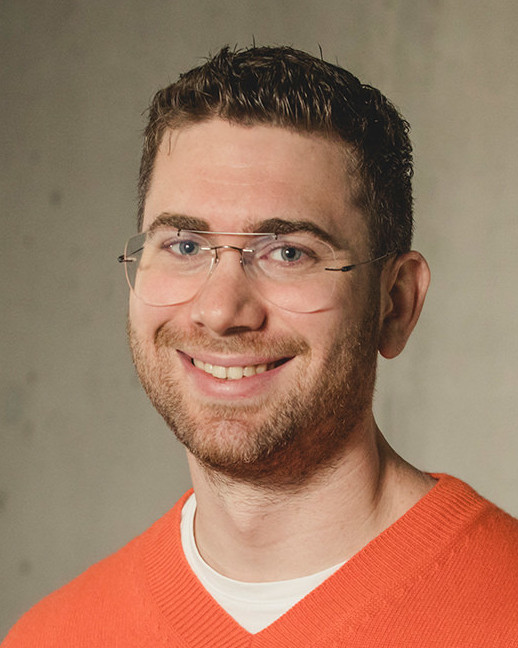}}]{Fabian Jenelten}
is a Ph.D. student at the Robotic Systems Lab, ETH Zurich, under the supervision of Prof. M. Hutter. His research interests are include model based and learning based control approaches for legged robot locomotion. He received his B.Sc. and M.Sc. in Mechanical Engineering from ETH Zurich, Switzerland, in 2015 and 2018.
\end{IEEEbiography}

%\vfill
%\newpage
%\vfill

\begin{IEEEbiography}[{\includegraphics[width=1in,height=1.25in,clip,keepaspectratio]{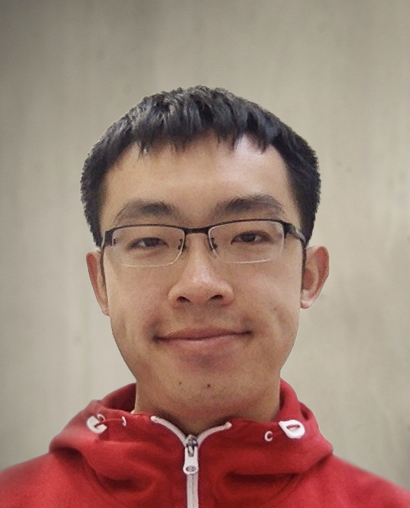}}]{Shaohui Yang}
obtained his B.Eng. in Computer Science from The Hong Kong University of Science and Technology (HKUST), China, in 2019, and his M.Sc. in Systems, Control and Robotics from KTH Royal Institute of Technology, Sweden, in 2022. After conducting his master thesis at the Robotic Systems Lab at ETH Zurich, he joined the Automatic Control Laboratory at EPFL as doctoral assistant, under the supervision of Prof. Colin Jones. His research interests lie in optimization, model predictive control and learning-based control. 
\end{IEEEbiography}

%\newpage

\begin{IEEEbiography}[{\includegraphics[width=1in,height=1.25in,clip,keepaspectratio]{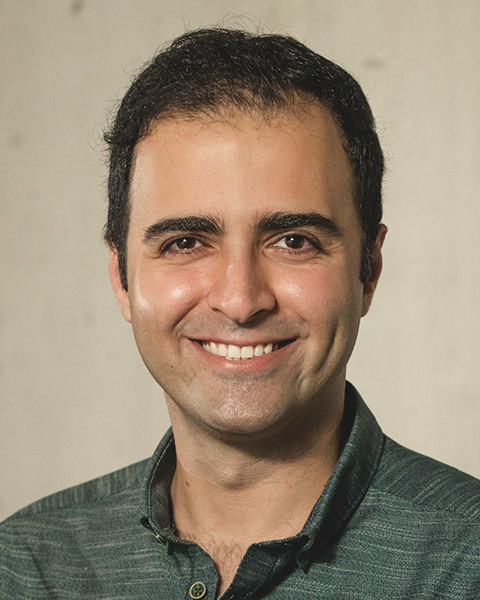}}]{Farbod Farshidian} is a Senior Scientist at Robotic System Lab, ETH Zurich. 
He received his M.Sc. in electrical engineering from the University of Tehran in 2012 and his Ph.D. from ETH Zurich in 2017. His research focuses on the motion planning and control of mobile robots, intending to develop algorithms and techniques to endow these robotic platforms to operate autonomously in real-world applications. Farbod is part of the National Centre of Competence in Research (NCCR) Robotics and NCCR Digital Fabrication.
\end{IEEEbiography}

\begin{IEEEbiography}[{\includegraphics[width=1in,height=1.25in,clip,keepaspectratio]{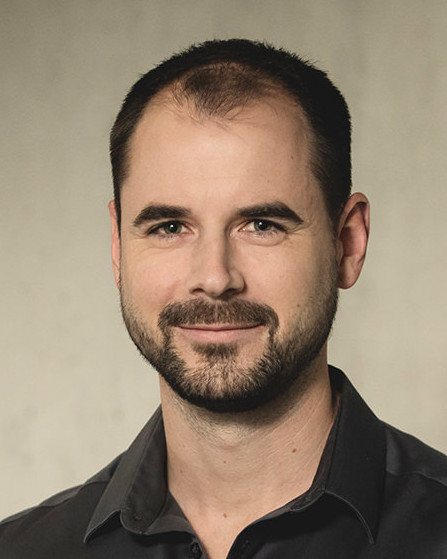}}]{Marco Hutter}
is Associate Professor for Robotic
Systems at ETH Zurich. He received his M.Sc. and PhD from ETH Zurich in 2009 and 2013. His research interests are in the development of novel machines and actuation concepts together with the underlying control, planning, and machine learning algorithms for locomotion and manipulation. Marco is part of the National Centre of Competence in Research (NCCR) Robotics and NCCR Digital Fabrication and PI in various international projects (e.g. EU NI) and challenges.
\end{IEEEbiography}

\vfill


%\vfill

%\enlargethispage{-2.2in}
